\documentclass[10pt,twocolumn,letterpaper]{article}

\usepackage{iccv}
\usepackage{times}
\usepackage{epsfig}
\usepackage{graphicx}
\usepackage{amsmath}
\usepackage{amssymb}

\usepackage{booktabs}       
\usepackage{nicefrac}       
\usepackage{xcolor}
\usepackage[binary-units=true]{siunitx}
\usepackage{bm}
\usepackage{import}
\usepackage{comment}
\usepackage{graphicx,wrapfig,lipsum}
\usepackage{subcaption}
\usepackage{transparent}

\DeclareMathOperator*{\argmin}{arg\,min}

\newcommand{\fs}{\bm{s}}
\newcommand{\fp}{\bm{p}}
\newcommand{\fx}{\bm{x}}

\newcommand{\rs}{\mathrm{s}}
\newcommand{\rp}{\mathrm{p}}
\newcommand{\norm}[1]{\left\|#1\right\|}

\usepackage[pagebackref=true,breaklinks=true,letterpaper=true,colorlinks,bookmarks=false]{hyperref}

\usepackage[breaklinks=true,bookmarks=false]{hyperref}
\iccvfinalcopy 

\def\iccvPaperID{****} 

\ificcvfinal\fi

\begin{document}

\newcommand{\OUR}{NPM}
\newcommand{\OURS}{NPMs}
\newcommand{\OURSFULL}{Neural Parametric Models}

\definecolor{abcolor}{RGB}{150,150,50}
\newcommand\AB[1] {\emph{\textcolor{abcolor}{AB: #1}}}

\definecolor{mzcolor}{RGB}{0,0,255}
\newcommand\MZ[1] {\emph{\textcolor{mzcolor}{MZ: #1}}}

\definecolor{jtcolor}{RGB}{255,0,255}
\newcommand\JT[1] {\emph{\textcolor{jtcolor}{JT: #1}}}

\definecolor{adcolor}{RGB}{50,150,150}
\newcommand\AD[1] {\emph{\textcolor{adcolor}{AD: #1}}}

\newcommand{\MATTHIAS}[1]{{\bf\textcolor{red}{Matthias: #1}}}

\definecolor{ppcolor}{RGB}{50,150,0}
\newcommand{\PP}[1]{{\bf\textcolor{ppcolor}{PP: #1}}}

\title{NPMs: Neural Parametric Models for 3D Deformable Shapes}

\author{
Pablo Palafox$^1$~~~
Aljaž Božič$^1$~~~
Justus Thies$^{1,2}$~~~
Matthias Nie{\ss}ner$^1$~~~
Angela Dai$^1$~~~
\vspace{0.2cm} \\ 
$^1$Technical University of Munich~~~
$^2$Max Planck Institute for Intelligent Systems, Tübingen
\vspace{0.2cm}
}

\twocolumn[{
	\renewcommand\twocolumn[1][]{#1}%
	\maketitle
	\begin{center}
	    \vspace{-0.475cm}
		\fontsize{9pt}{11pt}\selectfont
        \def\svgwidth{\linewidth}
\begingroup%
  \makeatletter%
  \providecommand\color[2][]{%
    \errmessage{(Inkscape) Color is used for the text in Inkscape, but the package 'color.sty' is not loaded}%
    \renewcommand\color[2][]{}%
  }%
  \providecommand\transparent[1]{%
    \errmessage{(Inkscape) Transparency is used (non-zero) for the text in Inkscape, but the package 'transparent.sty' is not loaded}%
    \renewcommand\transparent[1]{}%
  }%
  \providecommand\rotatebox[2]{#2}%
  \newcommand*\fsize{\dimexpr\f@size pt\relax}%
  \newcommand*\lineheight[1]{\fontsize{\fsize}{#1\fsize}\selectfont}%
  \ifx\svgwidth\undefined%
    \setlength{\unitlength}{550.03759603bp}%
    \ifx\svgscale\undefined%
      \relax%
    \else%
      \setlength{\unitlength}{\unitlength * \real{\svgscale}}%
    \fi%
  \else%
    \setlength{\unitlength}{\svgwidth}%
  \fi%
  \global\let\svgwidth\undefined%
  \global\let\svgscale\undefined%
  \makeatother%
  \begin{picture}(1,0.22317384)%
    \lineheight{1}%
    \setlength\tabcolsep{0pt}%
    \put(0,0){\includegraphics[width=\unitlength,page=1]{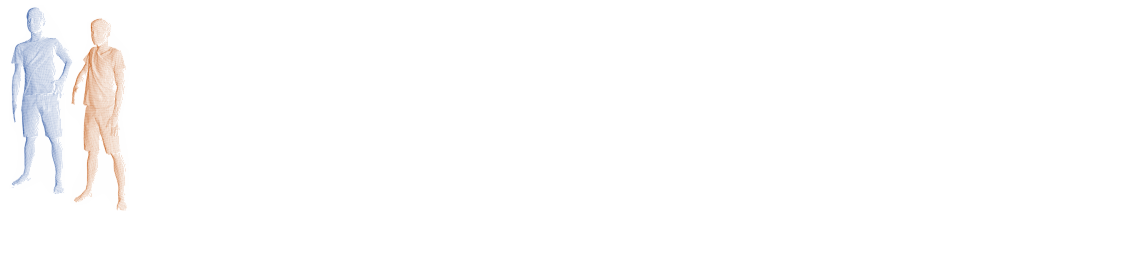}}%
    \put(0.02711627,0.04734012){\color[rgb]{0,0,0}\makebox(0,0)[lt]{\lineheight{1.25}\smash{\begin{tabular}[t]{l}$t_1$\end{tabular}}}}%
    \put(0.07936661,0.03553468){\color[rgb]{0,0,0}\makebox(0,0)[lt]{\lineheight{1.25}\smash{\begin{tabular}[t]{l}$t_2$\end{tabular}}}}%
    \put(0,0){\includegraphics[width=\unitlength,page=2]{teaser.pdf}}%
    \put(0.15692765,0.02583418){\color[rgb]{0,0,0}\makebox(0,0)[lt]{\lineheight{1.25}\smash{\begin{tabular}[t]{l}$t_N$\end{tabular}}}}%
    \put(0,0){\includegraphics[width=\unitlength,page=3]{teaser.pdf}}%
    \put(0.32848564,0.15990407){\color[rgb]{0,0,0}\makebox(0,0)[lt]{\lineheight{1.25}\smash{\begin{tabular}[t]{l}$\fs$\end{tabular}}}}%
    \put(0.0021992,0.00126964){\color[rgb]{0,0,0}\transparent{0.995}\makebox(0,0)[lt]{\lineheight{1.25}\smash{\begin{tabular}[t]{l}Monocular Depth Sequence\end{tabular}}}}%
    \put(0.24949671,0.19795915){\color[rgb]{0,0,0}\makebox(0,0)[t]{\lineheight{1.25}\smash{\begin{tabular}[t]{c}Latent \\Code \\Optimization\end{tabular}}}}%
    \put(0.42783607,0.00151289){\color[rgb]{0,0,0}\transparent{0.995}\makebox(0,0)[lt]{\lineheight{1.25}\smash{\begin{tabular}[t]{l}Shape\end{tabular}}}}%
    \put(0,0){\includegraphics[width=\unitlength,page=4]{teaser.pdf}}%
    \put(0.38224538,0.06698622){\color[rgb]{0,0,0}\makebox(0,0)[t]{\lineheight{1.25}\smash{\begin{tabular}[t]{c}Shape \\MLP\end{tabular}}}}%
    \put(0,0){\includegraphics[width=\unitlength,page=5]{teaser.pdf}}%
    \put(0.73747931,0.00010324){\color[rgb]{0,0,0}\transparent{0.995}\makebox(0,0)[lt]{\lineheight{1.25}\smash{\begin{tabular}[t]{l}Posed Reconstruction\end{tabular}}}}%
    \put(0.69301532,0.16117972){\color[rgb]{0,0,0}\makebox(0,0)[lt]{\lineheight{1.25}\smash{\begin{tabular}[t]{l}$\fp_N$\end{tabular}}}}%
    \put(0,0){\includegraphics[width=\unitlength,page=6]{teaser.pdf}}%
    \put(0.74158369,0.06709804){\color[rgb]{0,0,0}\makebox(0,0)[t]{\lineheight{1.25}\smash{\begin{tabular}[t]{c}Pose \\MLP\end{tabular}}}}%
    \put(0,0){\includegraphics[width=\unitlength,page=7]{teaser.pdf}}%
    \put(0.80863825,0.02749948){\color[rgb]{0,0,0}\makebox(0,0)[lt]{\lineheight{1.25}\smash{\begin{tabular}[t]{l}$t_1$\end{tabular}}}}%
    \put(0.87494018,0.02749948){\color[rgb]{0,0,0}\makebox(0,0)[lt]{\lineheight{1.25}\smash{\begin{tabular}[t]{l}$t_2$\end{tabular}}}}%
    \put(0.95207694,0.02749948){\color[rgb]{0,0,0}\makebox(0,0)[lt]{\lineheight{1.25}\smash{\begin{tabular}[t]{l}$t_N$\end{tabular}}}}%
    \put(0,0){\includegraphics[width=\unitlength,page=8]{teaser.pdf}}%
    \put(0.65826371,0.16117972){\color[rgb]{0,0,0}\makebox(0,0)[lt]{\lineheight{1.25}\smash{\begin{tabular}[t]{l}$\fp_2$\end{tabular}}}}%
    \put(0.63616131,0.16117972){\color[rgb]{0,0,0}\makebox(0,0)[lt]{\lineheight{1.25}\smash{\begin{tabular}[t]{l}$\fp_1$\end{tabular}}}}%
    \put(0,0){\includegraphics[width=\unitlength,page=9]{teaser.pdf}}%
  \end{picture}%
\endgroup%

	    \vspace{-0.35cm}
		
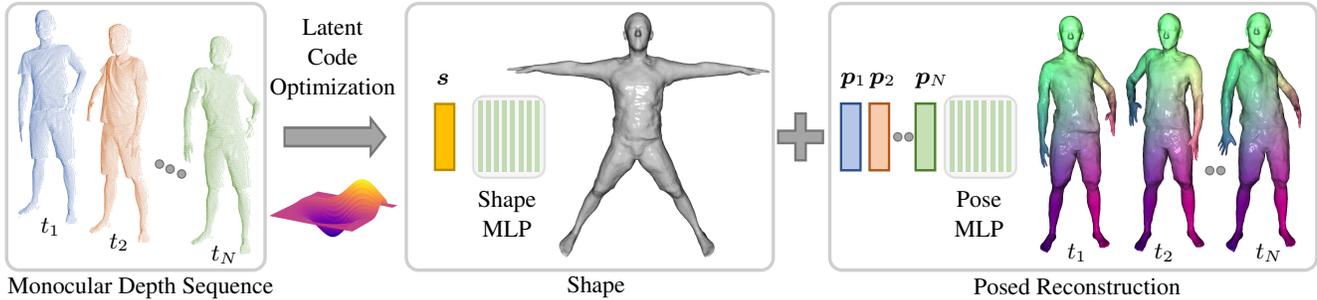
\captionof{figure}{
		Given an input monocular depth sequence, our \OURSFULL{} (\OURS{}), composed of learned latent shape and pose spaces, enable optimizing over the spaces to fit to the observations at test time, similar to traditional parametric model fitting (e.g., SMPL~\cite{loper2015smpl}). 
		\OURS{} can be constructed from a dataset of deforming shapes without strong requirements on surface correspondence annotations or category-specific knowledge.
		Our implicit shape and pose spaces enable expression of finer-scale details while providing a well-regularized space to fit to new observations of deforming shapes.}
		\label{fig:teaser}
		\vspace{-0.0775cm}
	\end{center}
}]

\ificcvfinal\thispagestyle{empty}\fi

\begin{abstract}
Parametric 3D models have enabled a wide variety of tasks in computer graphics and vision, such as modeling human bodies, faces, and hands.
However, the construction of these parametric models is often tedious, as it requires heavy manual tweaking, and they struggle to represent additional complexity and details such as wrinkles or clothing.
To this end, we propose \OURSFULL{} (\OURS{}), a novel, learned alternative to traditional, parametric 3D models, which does not require hand-crafted, object-specific constraints.
In particular, we learn to disentangle 4D dynamics into latent-space representations of shape and pose, leveraging the flexibility of recent developments in learned implicit functions.
Crucially, once learned, our neural parametric models of shape and pose enable optimization over the learned spaces to fit to new observations, similar to the fitting of a traditional parametric model, \eg, SMPL. 
This enables \OURS{} to achieve a significantly more accurate and detailed representation of observed deformable sequences.
We show that \OURS{} improve notably over both parametric and \mbox{non-parametric} state of the art in reconstruction and tracking of monocular depth sequences of clothed humans and hands.
Latent-space interpolation as well as shape~/~pose transfer experiments further demonstrate the usefulness of \OURS{}.
Code is publicly available at \href{https://pablopalafox.github.io/npms}{https://pablopalafox.github.io/npms}.
\end{abstract}
\section{Introduction}
%
Modeling deformable surfaces is fundamental towards understanding the 4D world that we live in, as well as creating or manipulating dynamic content.
While significant progress has been made in understanding the reconstruction of 3D shapes \cite{choy20163d,dai2017shape,fan2017point,wang2018pixel2mesh,park2019deepsdf,mescheder2019occupancyNet}, representing dynamic, deforming surfaces remains challenging.

Over the past years, parametric 3D models have seen remarkable success for domain-specific representations, such as for human bodies (\eg, SCAPE \cite{anguelov2005scape}, SMPL \cite{loper2015smpl}, Adam \cite{joo2018total}), hands (MANO \cite{MANO:SIGGRAPHASIA:2017}), animals (SMAL \cite{Zuffi_CVPR_2017}) and faces (\cite{paysan20093d}, FLAME \cite{li2017learning_flame}, \cite{ploumpis2019combining}).
These models have enabled a wide range of exciting applications and are instrumental in  modeling deformable 3D objects.
However, the construction of such a parametric model is a rather complex and tedious task, requiring notable manual intervention and incorporation of object-specific constraints in order for the parametric model to well-represent the space of possible shapes and deformations.
Moreover, such parametric models often struggle to represent additional complexity and details of deforming shapes, \eg, clothing, hair, \etc.

We propose \emph{\OURSFULL{}} (\OURS{}), an alternative formulation to traditional parametric 3D models where we learn a disentangled shape and pose representation that can be used like a traditional parametric model to fit to new observations.
We leverage the representation power of implicit functions to learn disentangled shape and pose spaces from a dataset that does not require  surface registration among all samples; this flexibility enables training on a wider variety of data.
We also do not make object-specific assumptions about the kinematic chain, the number of parts or the skeleton.
For training, our approach only requires that the same identity or shape can be seen in different poses, including a canonical pose.

Once trained, we can leverage our learned shape and pose representations as regularized spaces to be smoothly optimized over to fit to new observations at test time.
Additionally, our disentangled implicit representations of shape and pose enable modeling arbitrary connectivity and topology, as well as finer-scale levels of detail.
Thus, optimizing over our shape and pose spaces during inference enables representation of a globally consistent shape and temporally consistent poses while maintaining geometric fidelity.

Given a dataset of various shape identities and possibly different topologies, as well as various deformations of each shape identity (but without requiring registration of any identity to the others), we then train both shape and pose spaces in auto-decoder fashion.
We learn a shape code for each identity, with shape codes representing an SDF of the shape geometry.
Pose codes represent the flow field from the canonical shape of an identity to a given posed shape of the same identity.
Flow predictions are conditional on both a shape and a pose latent code, in order to represent shape-dependent deformations as well as to help learning disentangled shape and pose spaces. 

We demonstrate our \OURSFULL{} on the task of reconstruction and tracking of monocular depth sequences, as well as their capability in shape and pose transfer and interpolation.
In comparison with state-of-the-art parametric 3D models and implicit 4D representations, our \OURS{} capture higher-quality reconstructions with finer detail and more accurate non-rigid tracking.

In summary, we present the following key contributions:
\begin{itemize}
    \item We propose an alternative formulation to traditional parametric 3D deformable models, where shape and pose are disentangled into separate latent spaces via two feed-forward networks that are learned from data alone, \ie, without requiring domain-specific knowledge such as the  kinematic chain or number of parts.
    
    \item Importantly, our approach shows regularization capabilities that enable test-time optimization over the latent spaces of shape and pose to tackle the challenging task of fitting a model to monocular depth sequences, while retaining the detail present in the data.
\end{itemize}

\section{Related Work}

\paragraph{Traditional parametric models.}
Parametric 3D models have become a predominant approach to disentangle 3D deformable shapes into several factors, \eg, shape and pose, for domains such as human bodies \cite{anguelov2005scape, loper2015smpl, joo2018total, xu2020ghum}, hands \cite{MANO:SIGGRAPHASIA:2017}, animals \cite{Zuffi_CVPR_2017} and faces \cite{paysan20093d, li2017learning_flame, ploumpis2019combining}.
SMPL~\cite{loper2015smpl} is a very popular parametric model of the human body based on blend shapes in combination with a skeleton, and constructed from a dataset of 3D body scans.
Extensions exist to also model soft tissue \cite{pons2015dyna} and clothing \cite{bhatnagar2019multi, tiwari2020sizer, patel2020virtual, ma2020learning, pons2017clothcap, alldieck2019learning}.
To construct such parametric models, various domain-specific annotations are often required, such as the number of parts or the kinematic chain.
In contrast, our \OURS{} can be learned from data belonging to a domain without requiring any expert knowledge or manual intervention.
Additionally, approaches such as SMPL~\cite{loper2015smpl} or GHUM~\cite{xu2020ghum} are skinned vertex-based models which can struggle to represent complex surface features (e.g., wrinkles, clothing).
By leveraging recently proposed implicit functions, our approach can naturally capture more intricate surface detail.

\paragraph{Implicit representations for 3D shapes.}
Implicit representations such as Signed Distance Fields (\mbox{SDFs}) have been widely used to represent surfaces for 3D reconstruction, both static \cite{izadi2011kinectfusion,newcombe2011kinectfusion,niessner2013hashing,dai2017bundlefusion} and dynamic \cite{newcombe2015dynamicfusion, slavcheva2017killingfusion, bozic2020deepdeform}.
Various approaches to learn 3D shape generation have thus also leveraged such implicit definitions of a surface represented in a volumetric grid, where the regular structure is well-suited for convolutions but also induces cubic memory growth for high resolutions \cite{wu20153d,dai2017complete}.
 
Recent work in learning continuous implicit functions to represent shapes removes the explicit grid structure limitation, and shows strong promise in generating 3D shapes \cite{chen2019learning, genova2019learning, mescheder2019occupancyNet, michalkiewicz2019deep, park2019deepsdf, chibane2020implicit}.
In particular, DeepSDF~\cite{park2019deepsdf} proposes a feed-forward network to predict an SDF value given a query location conditional on a latent code that represents the shape, trained in auto-decoder fashion.
However, these approaches produce static surfaces that are not controllable, since shape and pose are entangled within a latent code.
Our approach leverages the representation power of implicit functions to learn disentangled implicit spaces --one for shape and one for pose--, enabling controllable 3D models which can be used to fit dynamic data or generate new posed shapes through interpolation in the spaces.

\paragraph{Learned representations for deformable shapes.} 
Recently, various learned approaches for representing deformable objects have been proposed \cite{groueix20183d, zhou2020unsupervised, niemeyer2019occupancyFlow, bhatnagar2020ipnet, bozic2020neuraltracking, bovzivc2020neural, li2020learning}.
Groueix \etal~\cite{groueix20183d} proposes a learned, template matching approach.
Zhou \etal~\cite{zhou2020unsupervised} learns disentangled shape and pose representations from datasets of registered meshes via a combination of self-consistency and cross-consistency constraints, without requiring expert knowledge in the dataset.
Our \OURS{} also do not require any manual annotations, but, unlike \cite{zhou2020unsupervised}, we do not require a template nor registration of identities in the dataset to each other.
This enables our \OURS{} to represent complex detail and broader shape variety, such as clothed bodies; moreover, we further explore how our learned latent spaces can be optimized over at test time to fit to sparse observations.

The recently proposed OFlow~\cite{niemeyer2019occupancyFlow} builds on the implicit 3D OccNet~\cite{mescheder2019occupancyNet} to learn 4D reconstruction from images or sparse point clouds.
OFlow learns a temporally and spatially continuous vector field which assigns a motion vector to every point in space and time, opening a promising avenue for spatio-temporal reconstruction, but remaining limited to very short sequences.
IP-Net~\cite{bhatnagar2020ipnet} presents an approach to combine learned implicit functions and traditional parametric models to produce controllable models of humans.
An implicit network \cite{chibane2020implicit} predicts inner body surface and outer detailed surface, to which SMPL+D~\cite{alldieck2019learning, lazova2019360} is fit for controllability.
\OURS{} also aim to provide a controllable model, but rather than using a SMPL basis, we learn a parametric model of disentangled shape and pose latent spaces which enable fitting by optimizing over the spaces jointly.
\section{Method}
We introduce \OURSFULL{} (\OURS{}), a learned approach to construct parametric 3D models from a dataset of different posed identities; unlike traditional parametric 3D models, we do not require the dataset to have annotations for domain-specific properties  such as the kinematic chain, skeleton or the surface-to-part mappings.
To construct our \OURS{}, we learn a latent space of (canonically-posed) shapes, along with a latent space of poses conditional on the shape.
We can then optimize jointly over the learned shape and pose spaces to fit to a new observation.

\begin{figure*}[t]
    \centering
    \fontsize{9pt}{11pt}\selectfont
    \def\svgwidth{\linewidth}
    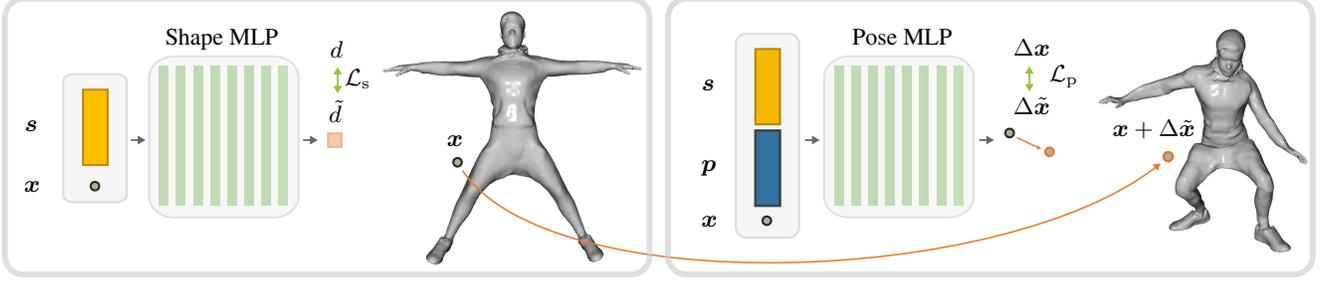
    \vspace{-0.3cm}
    \caption{
        \textbf{Architecture Overview}. To train our \OURS{}, we first learn a latent space of shape identities in their canonical poses (\eg, T-pose) by conditioning our shape MLP on the shape code $\fs_{i}$ assigned to each $i$-th identity. 
        Given this learned shape space, we learn a deformation field around the canonically-posed shape which maps points from this shape's canonical space to a $j$-th posed version of the shape. 
        We thus train a pose MLP conditioned on both the identity's latent shape code $\fs_{i}$ and the corresponding latent pose code $\fp_{j}$ to predict a flow vector $\Delta \fx$ for a query point $\fx$ sampled in the canonical pose.
    }
    \label{fig:architecture_overview}
\end{figure*}

Figure~\ref{fig:architecture_overview} shows an overview of our approach.
We employ an implicit representation for the shape space, encoding the SDF value for an input point, as well as for the pose space, encoding flow from the canonical to the deformed pose for an input point.
These implicit representations, coupled with the joint optimization over the learned spaces, enable capturing details present in the input data while effectively regularizing the shape and pose latent spaces.

\subsection{\OURSFULL{}} \label{sec:npm}
Given a dataset of meshes featuring a set of shape identities from the same class category in different poses, our goal is to learn a parametric model that not only regularizes the shape and pose latent spaces of the object class, but remains expressive enough to capture local details when fitting the learned model to new observations.
To learn \OURS{} from a dataset, the latter should follow two simple constraints: (1) each shape identity is posed canonically (\eg, T-pose), and (2) each shape identity has several posed or deformed instances for which dense surface correspondences to the canonical shape are available.
Various existing datasets (\eg, AMASS~\cite{AMASS:ICCV:2019}, DeformingThings4D~\cite{yang20204dcomplete}, CAPE~\cite{CAPE:CVPR:20}, MANO~\cite{MANO:SIGGRAPHASIA:2017}, etc.) easily fulfill these requirements.

We construct our \OURS{} by learning disentangled shape and pose spaces, leveraging implicit representations characterized by separate Multi-Layer Perceptrons (MLPs), one for shape and one for pose.
The shape encoding learns to implicitly represent the different identities in their canonical pose.
The pose space is conditional on both pose and shape codes, and learns a continuous deformation field around a canonical shape mapping points from this canonical shape to a deformed shape.

\subsection{Learned Shape Space}\label{sec:learned_shape_space}
Our shape space is learned by an MLP which predicts the implicit SDF for shape identities in their canonical pose; the shape is then defined as the zero iso-surface decision boundary and can be extracted with Marching Cubes~\cite{lorensen1987marching}.

Our shape MLP is trained in auto-decoder fashion as proposed by DeepSDF~\cite{park2019deepsdf}, where no encoder is used during training, directly optimizing over the latent code space.
Each canonically-posed shape identity $i$ in the training set is encoded in a $D_{\rs}$-dimensional latent shape code $\fs_i$.
The shape MLP learns to map an input point $\fx \in \mathbb{R}^3$ in the canonical space, conditioned on $\fs_i$, to an SDF value prediction $\tilde{d}$:
\begin{equation}
    f_{\theta_{\rs}} : \mathbb{R}^3 \times \mathbb{R}^{D_{\rs}} \rightarrow \mathbb{R}, \quad \left( \fs_i, \fx \right) \mapsto f_{\theta_{\rs}} \left( \fs_i, \fx \right) = \tilde{d}.
\end{equation}
We train our shape MLP on the $S$ shape identities of the dataset, in their canonical poses (see Fig.~\ref{fig:architecture_overview}).
To this end, we first normalize our training shapes (both canonically-posed and randomly posed) to reside within a unit bounding box by dividing all shapes by the extent of the largest bounding box in the dataset.
We then make our $S$ canonical shapes watertight. 
Note that the arbitrarily posed shapes used for training the pose MLP do not need to be watertight.

Next, for each $i$-th shape identity in the train set, we sample $N_{\rs}$ points $\{\fx_i^k\}_{k=1}^{N_{\rs}} \in \mathbb{R}^3$ along with their corresponding SDF values $\{d_i^k\}_{k=1}^{N_{\rs}} \in \mathbb{R}$.
These train samples come from two sources: (1)~$N_{\rs}^{\mathrm{ns}}$ near-surface points sampled randomly within a distance of \SI{0.05}{} from the surface of the shape and (2)~$N_{\rs}^{\mathrm{u}}$~points uniformly sampled within the unit bounding box, such that $N_{\rs} = N_{\rs}^{\mathrm{ns}} + N_{\rs}^{\mathrm{u}}$.
We refer readers to the appendix for further details.

Finally, to learn the latent shape space we minimize the following reconstruction energy over all shape identities in their canonical pose with respect to the individual shape codes $\{ \fs_i \}_{i=1}^S$ and the shape MLP weights $\theta_{\rs}$:
\begin{equation}
    \argmin_{\theta_{\rs}, \{ \fs_i \}_{i=1}^S} \sum_{i=1}^S \Big(  \sum_{k=1}^{N_{\rs}} \mathcal{L}_{\rs}(f_{\theta_{\rs}} (\fs_i, \fx_i^k), d_i^k) + \frac{\norm{\fs_i}^2_2}{\sigma_{\rs}^2} \Big),
\end{equation}
where $\mathcal{L}_s$ is a truncated $\ell_1$-loss on the predicted SDF $\tilde{d}^k_i$:
\begin{equation}
    \mathcal{L}_{\rs}(\tilde{d}^k_i, d^k_i) = \big| \, \mathrm{clamp}(\tilde{d}^k_i, \delta) - \mathrm{clamp}(d^k_i, \delta) \, \big|
    \label{eq:campled_L1}
\end{equation}
and $\mathrm{clamp}(d, \delta) := \min(\max(-\delta, d), \delta)$ defines the truncation region over which we maintain a metric SDF.
The $\ell_2$-regularization on the latent codes, controlled by the parameter $\sigma_{\rs}$, is required to enforce a compact shape manifold, as was found in \cite{park2019deepsdf}.

\vspace{-0.15cm}
\paragraph{Implementation details.} 
We use eight fully-connected, $F_{\rs}$-dimensional layers in our shape MLP, with ReLUs, and a final fully-connected layer followed by tanh which regresses the scalar SDF value. 
In our experiments, $F_{\rs}=512$ and $D_{\rs}=256$.
We use the Adam optimizer~\cite{kingma2014adam} and learning rates of \SI{5e-4}{} and \SI{1e-3}{} for the shape MLP $f_{\theta_{\rs}}$ and the shape codes $\{ \fs_i \}_{i=1}^{S}$, respectively.
Additionally, we apply a learning rate decay factor of $0.5$ every $500$ epochs.
We employ a regularization of $\sigma_s = 10^2$ on the shape codes, and set the SDF truncation to $\delta=0.1$.
The latent shape codes $\{ \fs_i \}_{i=1}^S$ are initialized randomly from $\mathcal{N}(0,\,0.01^{2})$.

\subsection{Learned Pose Space} \label{sec:learned_pose_space}
Our pose space is learned by an MLP which predicts a deformation field $f_{\theta_{\rp}}$ that maps points around identities in their canonical pose to the corresponding point locations in the space of the deformed pose. 
In particular, for a query point $\fx$ in the canonical space of identity $i$, the pose MLP predicts a flow vector $\Delta \tilde{\fx}$ that deforms the point from the canonical to the deformed space $j$, conditional on a \mbox{$D_{\rp}$-dimensional} latent pose code $\fp_j$ as well as on the latent shape code $\fs_i$.
This flow prediction is conditional on both $\fs_i$ and $\fp_j$, since the flow for a deformed pose $j$ of a given identity $i$ will depend on the shape itself (\eg, the flow to the same semantic pose for a large person vs.\ a small person will look different).
Formally, we have:
\begin{align*}
    f_{\theta_{p}} : \mathbb{R}^3 \times \mathbb{R}^{D_{\rs}} \times \mathbb{R}^{D_{\rp}} &\rightarrow \mathbb{R}^3, \\
    \left( \fs_i, \fp_j, \fx \right) &\mapsto f_{\theta_{p}} \left( \fs_i, \fp_j, \fx \right) = \Delta \tilde{\fx}.
\end{align*}
The pose MLP is trained on a set of $P$ deformation fields from the identities' canonical pose to the arbitrary poses available for each train identity.
Note that we do not require seeing every identity in the same pose, nor do we require an equal number of posed shapes for each identity.

For training, we sample $N_{\rp}$ surface points $\{ \fx_i^k \}_{k=1}^{N_{\rp}}$ on the previously normalized canonical shapes (see Sec.~\ref{sec:learned_shape_space}) for each \mbox{$i$-th} identity in the dataset; we also store the barycentric weights for each sampled point.
Each point is then randomly displaced a small distance $\bm{\delta n}$ along the normal direction of the corresponding triangle in the mesh.
Then, for each \mbox{$j$-th} posed shape available for the identity, we compute corresponding points $\{ \fx_j^{k} \}_{k=1}^{N_{\rp}}$ in the posed shape by using the same barycentric weights and $\bm{\delta n}$ to sample the posed mesh.
This approach gives us a deformation field (defined near the surface) between the canonical pose of a given identity $i$ and a deformed pose $j$ of the same identity.
For further sampling details, we refer to the appendix.

We use the ground truth flow vector $\Delta \fx_{ij}^k = \fx_j^k - \fx_i^k$ and define an $\ell_2$-loss $\mathcal{L}_{\rp}$ on the flow prediction $\Delta \tilde{\fx}_{ij}^k$.
To learn the pose space, we minimize the following energy over all $P$ deformation fields with respect to the individual pose codes $\{ \fp_j \}_{j=1}^P$ and pose MLP weights $\theta_{\rp}$:
\begin{equation}
    \argmin_{\theta_{\rp}, \{ \fp_j \}_{j=1}^P}
    \sum_{\substack{j=1 \\ i=m[j]}}^P \Big(  \sum_{k=1}^{N_{\rp}} \mathcal{L}_{\rp}(f_{\theta_{\rp}} (\fs_i, \fp_j, \fx_i^k), \Delta \fx_{ij}^k) + \frac{\norm{\fp_j}^2_2}{\sigma_{\rp}^2} \Big),
\end{equation}
where $m[\cdot]$ is a dictionary mapping the index $j$ of a posed shape to the corresponding index $i$ of its canonical shape, and $\sigma_{\rp}$ a regularization parameter for the pose codes. 
Note that we do not optimize over latent shape codes $\fs_i$ when learning the pose space.
However, we found that conditioning the pose MLP prediction on the latent shape codes was required to disentangle pose from shape.

\paragraph{Implementation details.}
Similar to the shape MLP, we use eight fully-connected, \mbox{$F_{\rp}$-dimensional} layers in our pose MLP, with ReLUs, followed by a final layer regressing the 3-dimensional flow vector $\Delta \tilde{\fx}$.
In our experiments, we use $F_{\rp}=1024$ and $D_{\rp}=256$.
We use the same training scheme as with the shape space training.

\subsection{Inference-time Optimization} \label{sec:inference_time_optim}
Once our latent representations of shape and pose have been constructed, we can leverage these spaces at test time by traversing them to solve for the latent codes that best explain an input sequence of $L$ depth maps.
We thus fit \OURS{} to the input data by solving for the unique latent shape code and the $L$ per-frame latent pose codes that best explain the whole sequence of observations.

For each depth map in the input sequence, we project the depth values into a $256^3$-SDF grid.
We also compute a volumetric mask $M_{\mathrm{o}}$ for occluded regions that are further than  \SI{0.01}{} (in normalized units) from the input observed surface, \ie, we do not consider grid points $\bm{g}$ where $\mathrm{SDF}(\bm{g}) < -0.01$.

We then obtain initial estimates of the shape code and pose codes with the initialization process described in Sec.~\ref{subsubsec:encoder_init}.
Given the initial shape code, we can extract the canonical shape surface and then sample $N_t$ surface points~$\{ \fx_k \}_{k=1}^{N_t}$ (\mbox{$N_t = \SI{500000}{}$} in our experiments) and add random displacements sampled from $\mathcal{N}(0,\,0.015^2)$.

To fit an \OUR{} to the monocular depth sequence, we minimize the following energy:
\begin{equation}
    \tilde{\fs}, \{ \tilde{\fp}_j \}_{j=1}^L = \argmin_{\fs, \{ \fp_j \}_{j=1}^L} \sum_{j=1}^L \sum_{\forall \fx_k} \mathcal{L}_{\mathrm{r}} + \mathcal{L}_{\mathrm{c}} + \mathcal{L}_{\mathrm{t}} + \mathcal{L}_{\mathrm{icp}}.
    \label{eq:test_time_optim}
\end{equation}
We use the same clamped $\ell_1$-loss from Eq.~\ref{eq:campled_L1} to define the reconstruction loss $\mathcal{L}_{\mathrm{r}}$:
\begin{equation}
    \mathcal{L}_{\mathrm{r}} = M_{\mathrm{o}} \; \mathcal{L}_{\rs} \Big( f_{\theta_{\rs}}(\fs, \fx_k), \left[ \fx_k + f_{\theta_{\rp}}(\fs, \fp_j, \fx_k) \right]_{\mathrm{sdf}} \Big),
\end{equation}
where $\left[ \cdot \right]_{\mathrm{sdf}}$ denotes trilinear interpolation of the SDF grid and $M_{\mathrm{o}}$ is the previously defined mask for occluded regions. 
Similar to train time, we enforce shape and pose code regularization:
\begin{equation}
    \mathcal{L}_{c} = \frac{1}{\sigma_{\rs}^2} \norm{\fs}^2_2 + \frac{1}{\sigma_{\rp}^2} \norm{\fp_j}^2_2,
\end{equation}
with $\sigma_{\rs}=10^{-1}$ and $\sigma_{\rp}=10^{-4}$.
Additionally, we enforce temporal regularization between the current frame $j$ and its neighboring frames \mbox{$Q = \{ j-1, j+1 \}$}. This is enforced with an $\ell_2$-loss on the pose MLP flow predictions for points $\fx_k$, and controlled with a weight of $\lambda_t = 200$:
\begin{equation}
    \mathcal{L}_{t} = \lambda_t \sum_{q \in Q} \norm{ f_{\theta_{\rp}}(\fs, \fp_j, \fx_k) - f_{\theta_{\rp}}(\fs, \fp_q, \fx_k) }^2_2.
\end{equation}
Finally, we employ an ICP-like loss $\mathcal{L}_{\mathrm{icp}}$ to further robustify the fitting (please refer to the appendix for details.)
We use the Adam optimizer~\cite{kingma2014adam} and learning rates of \SI{5e-4}{} and \SI{1e-3}{} for the shape and pose codes, respectively.

Given the optimized shape code and $L$ pose codes, to reconstruct the input sequence our approach only requires extracting the canonically-posed implicit surface via Marching Cubes \cite{lorensen1987marching} once (see Sec.~\ref{sec:learned_shape_space}).
We then deform the reconstructed canonical mesh into every frame in the input sequence by querying the pose MLP $f_{\theta_{\rp}}$ for every vertex in the canonical mesh.

\subsubsection{Predicting Shape and Pose Initializations}
\label{subsubsec:encoder_init}
To provide a good initialization for our latent code optimization, we train two 3D convolutional encoders $f_{\Omega_{\rs}}$ and $f_{\Omega_{\rp}}$ to predict initial estimates of the latent shape and pose codes, respectively.
Both encoders take as input the back-projected depth observation in the form of a partial voxel grid.
We then employ 3D convolutions and a final fully-connected layer to output a latent code estimate.
To train these encoders, we make use of the latent shape and pose vectors learned from the train set, and use them as target codes for training the encoders.
We found that this learned initialization provides robust initial code estimates, resulting in accurate reconstruction and tracking results.
Additional architecture details can be found in the appendix.

\section{Experiments}

We evaluate our \OURS{} on both synthetic and real-world datasets on the task of model fitting to a monocular depth sequence observation (Sec.~\ref{sec:model_fitting_experiments}).
We additionally demonstrate shape and pose transfer in Sec.~\ref{sec:transfer_results}, and show that our learned shape and pose spaces exhibit smooth, clear interpolation through the spaces in Sec~\ref{sec:interpolation_results}.

\paragraph{Datasets.}

\OURS{} can be learned for any class of non-rigidly deformable objects. We perform a comprehensive comparison with state-of-the-art methods on clothed human datasets, and show the general applicability of our method by learning an \OUR{} for hands.
For clothed humans, we evaluate on the recent CAPE \cite{CAPE:CVPR:20} dataset, which provides real-world scans of clothed humans and their corresponding SMPL+D registration. 
We also demonstrate our approach on synthetic human-like identities from the \mbox{DeformingThings4D~\cite{yang20204dcomplete}} dataset.
For training, we use 45k arbitrarily posed shapes from 118 distinct identities: 33~from~\cite{yang20204dcomplete}, 35~from~\cite{CAPE:CVPR:20} (13 in different clothing), and 50~from~AMASS~\cite{AMASS:ICCV:2019}.
We test our human \OUR{} on 4 identities from \cite{CAPE:CVPR:20} and 4 from \cite{yang20204dcomplete}, on a total of over 1600 frames distributed across 8 sequences (4 from each dataset).
We also learn a hand \OUR{} from 40k posed shapes from 400 distinct identities from MANO~\cite{MANO:SIGGRAPHASIA:2017}, and test it on 500 frames, with 5 identities and sequences of 100 frames each.

\paragraph{Evaluation metrics.} 
We measure both  reconstruction and tracking performance.
To quantitatively measure reconstruction quality we report two established metrics (following the evaluation protocol of \cite{mescheder2019occupancyNet}), which are computed on a per-frame basis. 
\mbox{\emph{Intersection over union} (IoU)} measures the overlap between the predicted mesh and the groundtruth mesh. 
We randomly sample $10^6$ points from the unit bounding box (where our normalized meshes reside) and determine if the points lie inside or outside the ground truth / predicted mesh.
\mbox{\emph{Chamfer}-$\ell_2$ (C-$\ell_2$)} offers a measure combining the accuracy and completeness of the reconstructed surface.
Following \cite{mescheder2019occupancyNet}, we use 100k randomly sampled surface points on the ground truth and predicted meshes.
Additionally, we evaluate tracking with \mbox{\emph{End-Point Error} (EPE)}, measuring the average $\ell_2$-distance between estimated keyframe-to-frame deformations and ground truth deformations as proposed in~\cite{bovzivc2020neural}; we sample 100k surface points and select a keyframe every 50 frames.

\subsection{Model Fitting to Monocular Depth Sequences}
\label{sec:model_fitting_experiments}

\begin{figure*}
    \centering
    \fontsize{9pt}{11pt}\selectfont
    \def\svgwidth{\linewidth}
    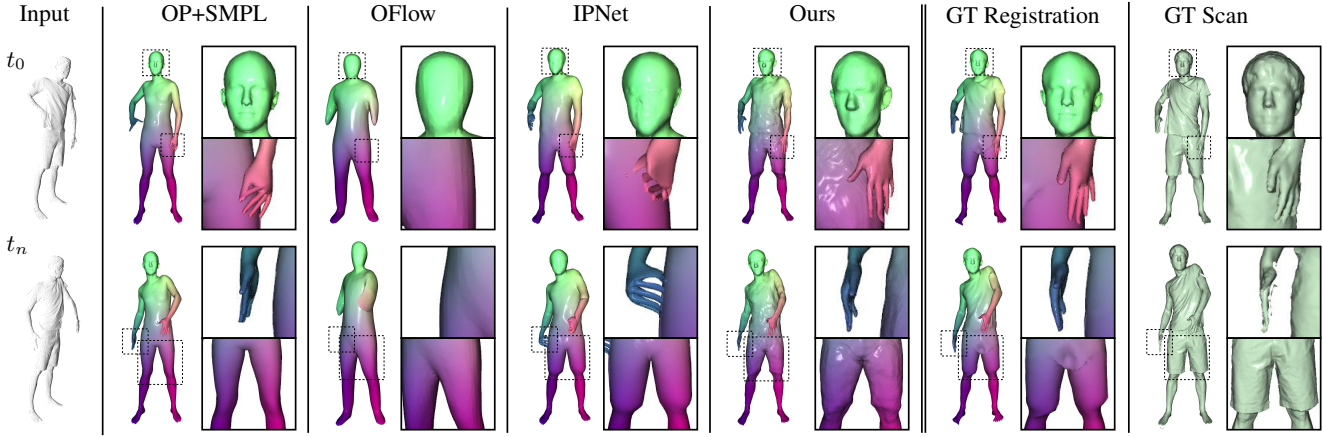
    \caption{Comparison to state-of-the-art methods on the task of model fitting to a monocular depth  sequence input (left column). From left to right, we compare with OpenPose~\cite{cao2019openpose} + SMPL~\cite{loper2015smpl}, OFlow~\cite{niemeyer2019occupancyFlow} and IP-Net~\cite{bhatnagar2020ipnet}; our \OURS{} effectively capture local details present in the input views. The last two columns show the ground truth registration provided by  CAPE~\cite{CAPE:CVPR:20} and the original scans, from which the input depth maps are rendered.} 
    \label{fig:qualitative_comparison_cape}
\end{figure*}

\paragraph{Real human data.}
We show a comparison to state of the art on monocular depth data rendered from CAPE~\cite{CAPE:CVPR:20} real scans in Tab.~\ref{tab:quantitative_comparison_CAPE}, and qualitatively in Fig.~\ref{fig:qualitative_comparison_cape}.
We compare with \mbox{SMPL~\cite{loper2015smpl}}, a state-of-the-art traditional parametric model, as well as the state-of-the-art deep-learning-based approaches \mbox{OFlow~\cite{niemeyer2019occupancyFlow}} and \mbox{IP-Net~\cite{bhatnagar2020ipnet}}.
We fit a SMPL model to the input depth maps by minimizing the reconstruction loss between surface points and the SDF grid extracted from the depth map (see Sec.~\ref{sec:inference_time_optim}), enforcing surface points to lie at the zero-level set of the SDF grid.
To guide this SMPL fitting, we use \mbox{OpenPose~\cite{cao2019openpose}} to provide sparse keypoint correspondences, minimizing the reprojection error between projected SMPL joints and OpenPose predictions.
To increase robustness, we also constrain the 3D error between SMPL joints in 3D and back-projected (using the input depth map) OpenPose predictions.
Temporal regularization is applied by minimizing vertex-to-vertex distances between neighboring frames.

IP-Net is trained on the same combination of human data employed to learn our human \OUR{}.
Since OFlow was developed for continuous sequences of up to 17 frames (and we found performance to degrade noticeably for longer sequences), we  prepare a train dataset of over 200k frames that fulfills this requirement; at test time, we evaluate the average of 17-frame subsequences covering the full test sequence.
Our approach to learn shape and pose spaces \mbox{--enabling} latent code optimization for \mbox{fitting--} provides both effective shape and pose regularization over the manifolds, while capturing local details.
This results in more accurate reconstruction and tracking performance.

\begin{table}[t]
	\resizebox{\linewidth}{!}{%
    \centering
    \begin{tabular}{lcccc}
        \toprule
        \textbf{Method} & \textbf{IoU} $\uparrow$ & \textbf{C-$\ell_2$} ($\times 10^{-3}$) $\downarrow$ & \textbf{EPE} ($\times 10^{-2}$) $\downarrow$ \\ 
        \midrule
        OpenPose+SMPL         & 0.68  & 0.243 & 2.82 \\
        OFlow*          & 0.55  & 0.755	& 2.65 \\
        IP-Net           & 0.82  & 0.034 & 2.52 \\
        \midrule
        Ours (no Shape Enc) & 0.83 & 0.023 & 0.77 \\
        Ours (no Pose Enc) & 0.78 & 0.174 & 3.61 \\
        Ours (no S\&P Enc) & 0.77 & 0.185 & 3.65 \\
        Ours               & \textbf{0.83} & \textbf{0.022} & \textbf{0.74} \\
        \bottomrule
    \end{tabular}
    }
    \caption{
        Comparison with state-of-the-art methods on real scans of CAPE \cite{CAPE:CVPR:20}.
        *Since OFlow \cite{niemeyer2019occupancyFlow} works only on sequences of up to 17 frames, we report the average over sub-sequences of such length.
        }
    \label{tab:quantitative_comparison_CAPE}
\end{table}

\begin{figure*}
    \centering
    \fontsize{9pt}{11pt}\selectfont
    \def\svgwidth{\linewidth}
    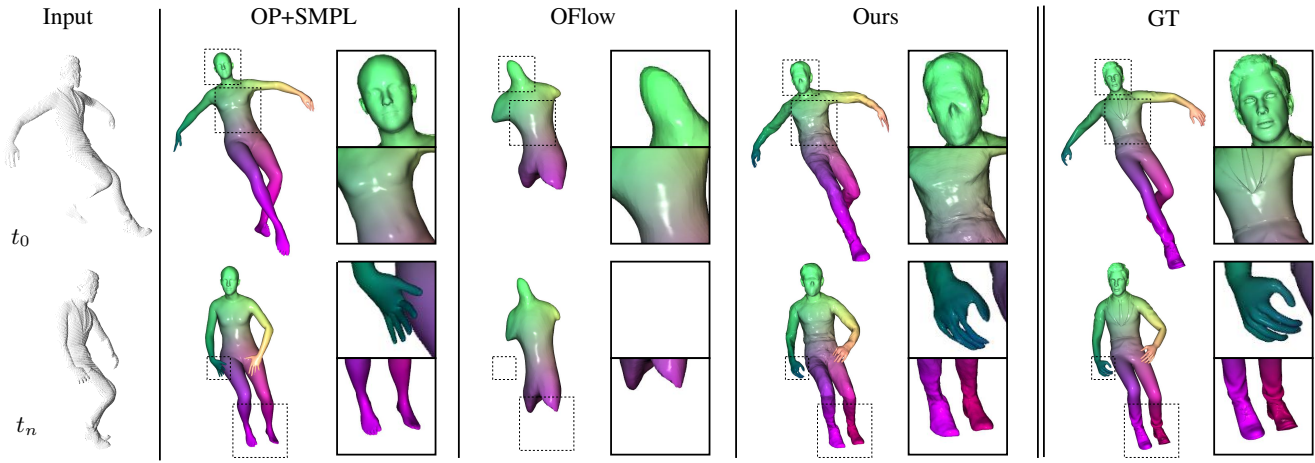
    \caption{Comparison to state-of-the-art methods on the task of model fitting to a monocular depth sequence input (left column) from a synthetic dataset (DeformingThings4D~\cite{yang20204dcomplete}). From left to right, we compare with OpenPose~\cite{cao2019openpose} + SMPL~\cite{loper2015smpl} and OFlow~\cite{niemeyer2019occupancyFlow}; our \OURS{} effectively capture local details present in the input views.} 
    \label{fig:qualitative_comparison_mixamo}
\end{figure*}

\vspace{-0.1cm}
\paragraph{Synthetic human data.}
We also evaluate on synthetic sequences from the DeformingThings4D~\cite{yang20204dcomplete} dataset, in comparison with SMPL~\cite{loper2015smpl} and OFlow~\cite{niemeyer2019occupancyFlow} in Tab.~\ref{tab:quantitative_comparison_MIXAMO}.
Our learned shape and pose spaces effectively capture significantly improved reconstruction and tracking in our model fitting experiments.
Fig.~\ref{fig:qualitative_comparison_mixamo} shows a qualitative comparison to state-of-the-art methods, demonstrating our global reconstruction and tracking along with the captured local detail.

\begin{table}[t]
	\resizebox{\linewidth}{!}{%
    \centering
    \begin{tabular}{lcccc}
        \toprule
        \textbf{Method} & \textbf{IoU} $\uparrow$ & \textbf{C-$\ell_2$} ($\times 10^{-3}$) $\downarrow$ & \textbf{EPE} ($\times 10^{-2}$) $\downarrow$ \\ 
        \midrule
        OpenPose+SMPL       & 0.64 & 0.251 & 2.04 \\
        OFlow*        & 0.40 & 2.688 & 7.52 \\
        \midrule
        Ours & \textbf{0.78} & \textbf{0.051} & \textbf{1.07} \\ 
        \bottomrule
    \end{tabular}
    }
    \caption{
        Comparison with state of the art on test sequences from DeformingThings4D~\cite{yang20204dcomplete}. 
        *Since OFlow \cite{niemeyer2019occupancyFlow} works only on sequences of up to 17 frames, we report the average over sub-sequences of such length.
    }
    \label{tab:quantitative_comparison_MIXAMO} 
    \vspace{-0.1cm}
\end{table}

\vspace{-0.1cm}
\paragraph{4D point cloud completion on D-FAUST.} 
We additionally compare with OFlow~\cite{niemeyer2019occupancyFlow} on their 4D point cloud completion task on D-FAUST~\cite{bogo2017dynamicFAUST} in Tab.~\ref{tab:quantitative_comparison_DFAUST}. 
We use a pretrained OFlow model provided by the authors and test on 20k dense point cloud trajectories sampled from the ground truth meshes.
For our method, we only consider a monocular sequence of depth maps as input, resulting in more partial observations.
Even with more partial data, our \OUR{} fitting achieves significantly improved performance.
\begin{table}[t]
    \centering
    \begin{tabular}{lcccc}
        \toprule
        \textbf{Method} & \textbf{IoU} $\uparrow$ & \textbf{C-$\ell_2$} ($\times 10^{-3}$) $\downarrow$ & \textbf{EPE} ($\times 10^{-2}$) $\downarrow$ \\ 
        \midrule
        OFlow & 0.74 & 0.105 &1.12 \\
        Ours & {\bf 0.83} & {\bf 0.019} & {\bf 0.61} \\
        \bottomrule
    \end{tabular}
    \vspace{-0.1cm}
    \caption{
        Comparison with OFlow~\cite{niemeyer2019occupancyFlow} on D-FAUST~\cite{bogo2017dynamicFAUST}.
    }
    \label{tab:quantitative_comparison_DFAUST}
    \vspace{-0.2cm}
\end{table}

\paragraph{What is the effect of the encoder initialization?}
In Table~\ref{tab:quantitative_comparison_CAPE}, we evaluate the effect of our encoder initialization for our \OURS{} optimization.
In place of the encoder predicted initialization, we use the average shape and pose latent codes from the train set.
We measure the effect of not using the shape encoder (\emph{no Shape Enc}), not using the pose encoder (\emph{no Pose Enc}), and not using neither the shape nor the pose encoder (\emph{no S\&P Enc}) for code initialization.
The shape and pose code estimates provided by our encoders result in a closer initialization and an improvement in reconstruction and tracking performance.
\vspace{-0.1cm}

\begin{figure}[t]
    \centering
    \fontsize{9pt}{11pt}\selectfont
    \def\gwidth{\linewidth}
    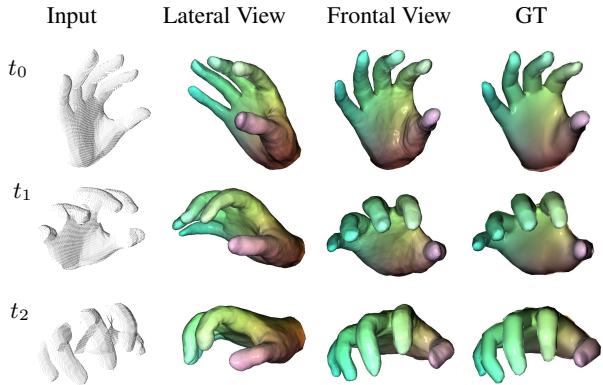
    \vspace{-0.5cm}
    \caption{Registration of our hand \OUR{} to a test sequence of monocular depth views generated using MANO \cite{MANO:SIGGRAPHASIA:2017}. 
    }
    \label{fig:model_fitting_MANO}
\end{figure}

\paragraph{Hand registration.}
\OURS{} can be constructed on various datasets with posed identities. We demonstrate its applicability to hand data generated with the MANO~\cite{MANO:SIGGRAPHASIA:2017} parametric model.
Fig.~\ref{fig:model_fitting_MANO} shows our hand \OUR{} fitting a test monocular depth sequence.
We accurately capture both global structure and smaller-scale details (\eg, creasing from bent knuckles), achieving an IoU of $0.86$, \mbox{Chamfer-$\ell_2$} of $1.39\times 10^{-5}$, and EPE of $5.89\times 10^{-3}$.

\subsection{Shape and Pose Transfer}
\label{sec:transfer_results}
\begin{figure}[t]
    \centering
    \fontsize{9pt}{11pt}\selectfont
    \def\svgwidth{\linewidth}
\begingroup%
  \makeatletter%
  \providecommand\color[2][]{%
    \errmessage{(Inkscape) Color is used for the text in Inkscape, but the package 'color.sty' is not loaded}%
    \renewcommand\color[2][]{}%
  }%
  \providecommand\transparent[1]{%
    \errmessage{(Inkscape) Transparency is used (non-zero) for the text in Inkscape, but the package 'transparent.sty' is not loaded}%
    \renewcommand\transparent[1]{}%
  }%
  \providecommand\rotatebox[2]{#2}%
  \newcommand*\fsize{\dimexpr\f@size pt\relax}%
  \newcommand*\lineheight[1]{\fontsize{\fsize}{#1\fsize}\selectfont}%
  \ifx\svgwidth\undefined%
    \setlength{\unitlength}{156.84625178bp}%
    \ifx\svgscale\undefined%
      \relax%
    \else%
      \setlength{\unitlength}{\unitlength * \real{\svgscale}}%
    \fi%
  \else%
    \setlength{\unitlength}{\svgwidth}%
  \fi%
  \global\let\svgwidth\undefined%
  \global\let\svgscale\undefined%
  \makeatother%
  \begin{picture}(1,1.18667578)%
    \lineheight{1}%
    \setlength\tabcolsep{0pt}%
    \put(0,0){\includegraphics[width=\unitlength,page=1]{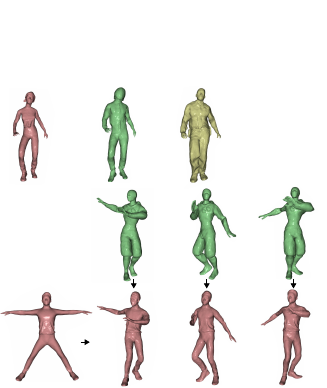}}%
    \put(0.00245635,1.02381629){\color[rgb]{0,0,0}\makebox(0,0)[lt]{\lineheight{1.25}\smash{\begin{tabular}[t]{l}Shape Transfer\end{tabular}}}}%
    \put(0,0){\includegraphics[width=\unitlength,page=2]{repose_reshape.pdf}}%
    \put(0.00514496,0.40392062){\color[rgb]{0,0,0}\makebox(0,0)[lt]{\lineheight{1.25}\smash{\begin{tabular}[t]{l}Pose Transfer\end{tabular}}}}%
  \end{picture}%
\endgroup%

    \vspace{-0.5cm}
    \caption{
        Shape and pose transfer with \OURS{}.
        We can transfer a given identity to a posed shape (shape transfer);  given a source identity in different poses, we can repose a target identity with the poses of the source (pose transfer).
    }
    \label{fig:pose_shape_transfer}
    \vspace{-0.2cm}
\end{figure}

\OURS{} enable shape and pose transfer: we can transfer a given identity to a posed shape (shape transfer), and given a source identity in different poses, we can repose a target identity with the poses of the source identity (pose transfer).
This is possible due to our disentangled shape and pose embedding spaces, which enables novel combinations of shape and pose latent codes.
In Fig.~\ref{fig:pose_shape_transfer} and the supplemental video, we show additional examples of shape and pose transfer.

\subsection{Latent-Space Interpolation}
\label{sec:interpolation_results}
\begin{figure}[t]
    \centering
    \fontsize{9pt}{11pt}\selectfont
    \def\svgwidth{\linewidth}
\begingroup%
  \makeatletter%
  \providecommand\color[2][]{%
    \errmessage{(Inkscape) Color is used for the text in Inkscape, but the package 'color.sty' is not loaded}%
    \renewcommand\color[2][]{}%
  }%
  \providecommand\transparent[1]{%
    \errmessage{(Inkscape) Transparency is used (non-zero) for the text in Inkscape, but the package 'transparent.sty' is not loaded}%
    \renewcommand\transparent[1]{}%
  }%
  \providecommand\rotatebox[2]{#2}%
  \newcommand*\fsize{\dimexpr\f@size pt\relax}%
  \newcommand*\lineheight[1]{\fontsize{\fsize}{#1\fsize}\selectfont}%
  \ifx\svgwidth\undefined%
    \setlength{\unitlength}{308.33238042bp}%
    \ifx\svgscale\undefined%
      \relax%
    \else%
      \setlength{\unitlength}{\unitlength * \real{\svgscale}}%
    \fi%
  \else%
    \setlength{\unitlength}{\svgwidth}%
  \fi%
  \global\let\svgwidth\undefined%
  \global\let\svgscale\undefined%
  \makeatother%
  \begin{picture}(1,0.78645819)%
    \lineheight{1}%
    \setlength\tabcolsep{0pt}%
    \put(0,0){\includegraphics[width=\unitlength,page=1]{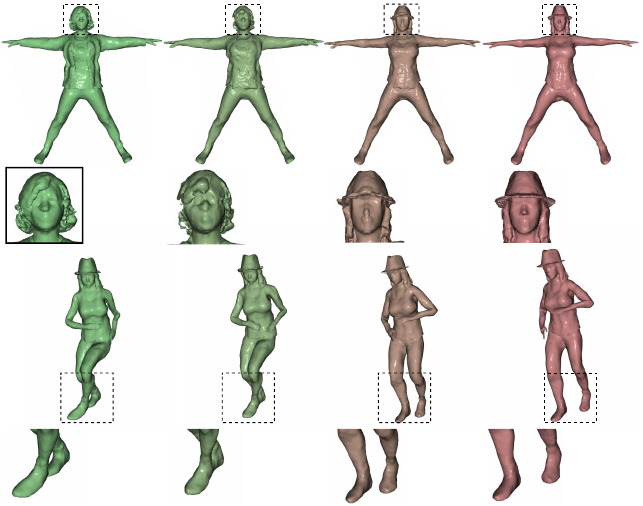}}%
    \put(0.03086635,0.58027234){\color[rgb]{0,0,0}\rotatebox{90}{\makebox(0,0)[lt]{\lineheight{1.25}\smash{\begin{tabular}[t]{l}Shape\end{tabular}}}}}%
    \put(0.03801333,0.21251599){\color[rgb]{0,0,0}\rotatebox{90}{\makebox(0,0)[lt]{\lineheight{1.25}\smash{\begin{tabular}[t]{l}Pose\end{tabular}}}}}%
    \put(0,0){\includegraphics[width=\unitlength,page=2]{interpolation.pdf}}%
  \end{picture}%
\endgroup%

    \caption{
        Shape and pose latent-space interpolation.
    }
    \label{fig:interpolation}
    \vspace{-0.15cm}
\end{figure}

Our latent spaces of shape and pose can be traversed to obtain novel shapes and poses.
Interpolation through the learned spaces (Fig.~\ref{fig:interpolation} and supplemental video) suggests continuity of our shape and pose latent spaces.

\vspace{-0.15cm}
\paragraph{Limitations.}
While \OURS{} demonstrate potential for constructing and fitting learned parametric models, several limitations remain.
For instance, our implicit representation of shape and pose deformation can struggle with very flat surfaces, as they comprise little volume and must have precisely defined inside~/~outside; incorporating semantic information into \OURS{} could help in this regard.
Although \OURS{} can capture fine-scale details present in the input data (e.g., clothing boundaries), high-frequency details (e.g., the outline of a tie) remain challenging.
Our learned spaces also do not consider the physics of deformation, which could encourage volume preservation and constrain deformations to physically correct movements.
\section{Conclusion}
In this paper, we introduced \OURSFULL{}, enabling the construction of learned parametric models with disentangled shape and pose representations which can accurately represent 4D sequences of dynamic objects.
Unlike traditional parametric models, our \OURS{} leverage learned implicit functions to expressively capture local details in shape and pose, and our test-time latent code optimization enables accurate fitting to observed details in input monocular depth sequences, outperforming parametric and learned 4D representations.
Our learned \OURS{} also enable effective shape and pose transfer, and demonstrate smooth interpolations within the spaces across new shapes and poses.
We additionally demonstrate more general applicability to a hands dataset, and believe this opens many promising avenues for various other domains in spatio-temporal modeling.

\section*{Acknowledgments}
This project is funded by the Bavarian State Ministry of Science and the Arts coordinated by the Bavarian Research Institute for Digital Transformation (bidt), a TUM-IAS Rudolf M\"o{\ss}bauer Fellowship, the ERC Starting Grant Scan2CAD (804724), and the German Research Foundation (DFG) Grant Making Machine Learning on Static and Dynamic 3D Data Practical.

\begin{appendix}
\section*{Appendix}
%
%
In this appendix, we provide additional details about our proposed \OURSFULL{}.
Specifically, we describe the training data preparation (Sec.~\ref{sec:data_preparation}), the network architectures (Sec.~\ref{sec:network_architecture}), as well as training details including the hyper-parameters (Sec.~\ref{sec:training_details}), and information about the inference-time optimization (Sec.~\ref{sec:inference_details}).
Additional qualitative evaluations and results are shown in the supplemental video.
 
\section{Data Preparation}
\label{sec:data_preparation}

\subsection{Waterproofing Canonical Shapes} \label{sec:waterproof}
We waterproof canonical shapes by first rendering depth maps of the non-watertight shape inside a virtual multi-view setup, then computing partial surface meshes from each depth map and finally running Poisson surface reconstruction \cite{kazhdan2006poisson} on the merged set of partial meshes.
Although this approach is not guaranteed to produce watertight meshes, especially when applied to meshes with large holes, we found it to work well on our canonical shapes, allowing us to retain details present in the original (non-watertight) meshes.

\subsection{Train Samples}

\begin{figure}
    \centering
    \includegraphics[width=\linewidth]{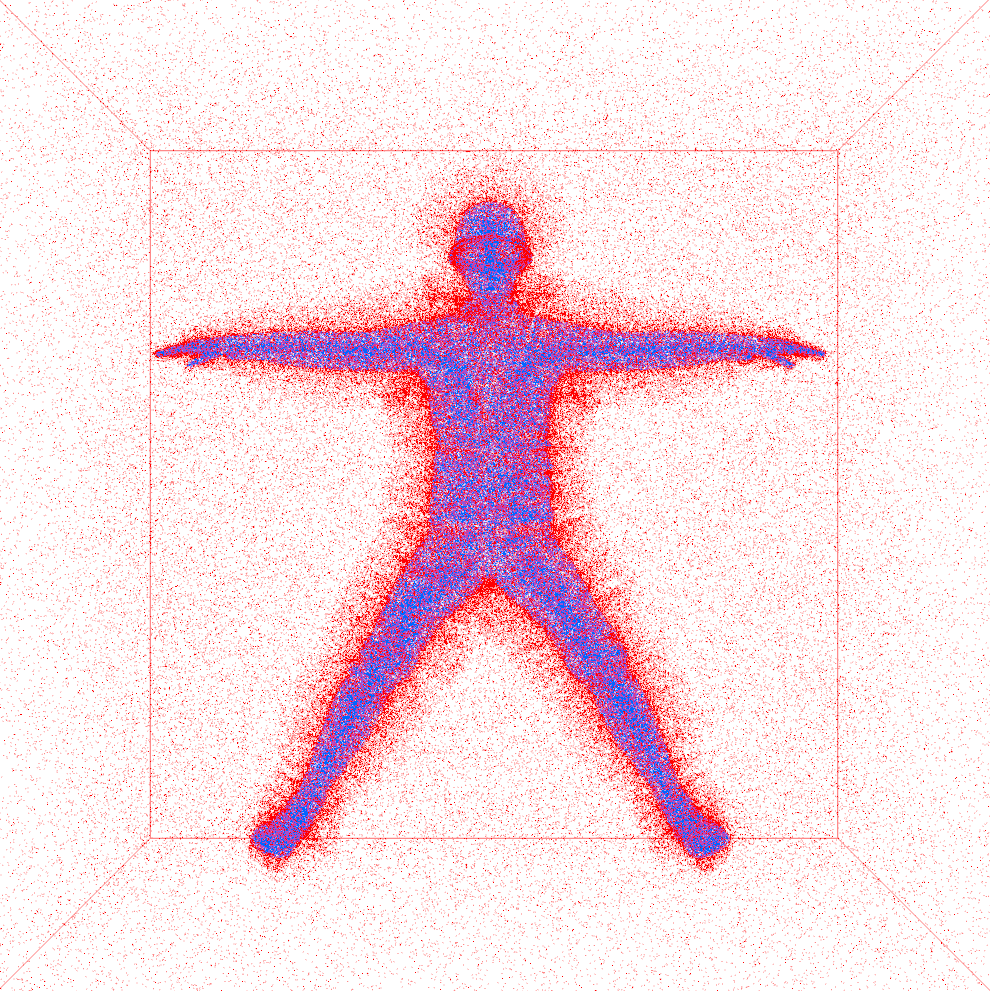}
    \caption{
    We visualize the $N_{\rs}$ training points available for a given identity for training the shape MLP.
    Points in red are outside of the canonical shape, and so have a positive SDF value.
    Points in blue, on the other hand, have a negative SDF value, since they reside within the shape.
    Note that we sample densely near the surface ($N_{\rs}^{\mathrm{ns}} = 300$k).
    Additionally, we sample points in the unit bounding box of the shape ($N_{\rs}^{\mathrm{u}} = 100$k).
    }
    \label{fig:sampling_visualization_SHAPE}
\end{figure}

\paragraph{Shape space.}
For each shape identity $i$ in the train dataset, given their watertight mesh in canonical pose, we sample a total number of $N_{\rs}$ points $\{ \fx_i^k \}_{k=1}^{N_{\rs}} \in \mathbb{R}^3$, along with their corresponding ground truth SDF value $\{ s_i^k \}_{k=1}^{N_{\rs}} \in \mathbb{R}$ (see Fig.~\ref{fig:sampling_visualization_SHAPE}).
As explained in the main paper, these samples come from two sources:
\begin{enumerate}
    \item $N_{\rs}^{\mathrm{ns}}$ near-surface points sampled randomly within a distance of \SI{0.05}{} from the surface of the shape.
    \item $N_{\rs}^{\mathrm{u}}$~points uniformly sampled within the unit bounding box, such that $N_{\rs} = N_{\rs}^{\mathrm{ns}} + N_{\rs}^{\mathrm{u}}$.
\end{enumerate}
In our experiments, we set $N_{\rs}^{\mathrm{ns}} = 300$k and $N_{\rs}^{\mathrm{u}} = 100$k, for a total of $N_{\rs} = 400$k.

%

\paragraph{Pose space.}
In Sec.~3.3 of the main text, we explain the procedure for generating flow samples for training the pose MLP; here, we provide additional details.
First, we sample $N_{\rp}$ surface points $\{ \fx_i^k \}_{k=1}^{N_{\rp}}$ on the normalized canonical shapes for each \mbox{$i$-th} identity in the dataset; we also store the barycentric weights for each sampled point.
Each point is then randomly displaced a small distance $\bm{\delta n} \sim \mathcal{N}(0,\,\bm{\Sigma}^2)$ along the normal direction of the corresponding triangle in the mesh, with $\bm{\Sigma} \in \mathbb{R}^3$ a diagonal covariance matrix with entries $\bm{\Sigma}_{ii} = \sigma$.
Then, for each \mbox{$j$-th} posed shape available for the identity, we compute corresponding points $\{ \fx_j^{k} \}_{k=1}^{N_{\rp}}$ in the posed shape by using the same barycentric weights and $\bm{\delta n}$ to sample the posed mesh.
This approach gives us a deformation field (defined near the surface) between the canonical pose of a given identity $i$ and a deformed pose $j$ of the same identity.

In our experiments, we pre-compute $N_{\rp} = 200$k correspondences for every \mbox{$j$-th} posed shape available in the dataset.
We obtain good results by sampling 50\% of these points with $\sigma = 0.01$, and 50\% with $\sigma = 0.002$ (in normalized coordinates).

\section{Network Architecture}
\label{sec:network_architecture}

\subsection{Shape and Pose Auto-decoders}

\begin{figure}[t]
    \centering
    \fontsize{7pt}{9pt}\selectfont
    \def\svgwidth{\linewidth}
    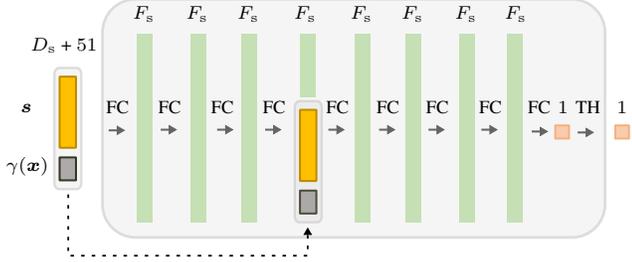
    \caption{
        A detailed visualization of our shape MLP.
    }
    \label{fig:shape_mlp_detail}
\end{figure}

\begin{figure}[t]
    \centering
    \fontsize{7pt}{9pt}\selectfont
    \def\svgwidth{\linewidth}
    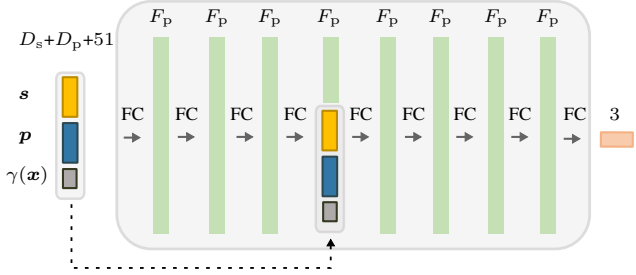
    \caption{
        A detailed visualization of our pose MLP.
    }
    \label{fig:pose_mlp_detail}
\end{figure}

Figures~\ref{fig:shape_mlp_detail} and \ref{fig:pose_mlp_detail} detail our feed-forward networks for learning our latent shape and pose spaces.
Both MLPs are composed of 8 fully connected layers, applied with weight-normalization \cite{salimans2016weight}.
We use ReLU activations as non-linearities after every intermediate layer.
After the final layer of our shape MLP we use tanh to regress an SDF value, whereas for our pose MLP we directly regress a 3-dimensional flow vector.
For both networks, a skip connection is used at the fourth layer.

To learn our human \OUR{}, we set the feature size $F_{\rs}$ of each layer in our shape MLP to $512$, and $D_{\rs} = 256$.
For learning a hand \OUR{}, we use $F_{\rs} = 64$ and $D_{\rs} = 16$.
When learning the latent pose space, we set \mbox{$F_{\rp} = 1024$} and \mbox{$D_{\rp} = 256$} in the case of humans, and $F_{\rp} = 256$ and $D_{\rp} = 64$ for hands.
We use the positional encoding proposed in \cite{mildenhall2020nerf} to encode the query point $\fx$ for both our shape and pose MLPs, and use $8$ frequency bands.
Positional encoding is denoted by $\gamma(\fx)$ in Figures~\ref{fig:shape_mlp_detail} and \ref{fig:pose_mlp_detail}.

\subsection{Shape and Pose Encoders for Initialization}
\label{sec:encoder_init_details}
As presented in the main paper, to provide a good initialization for our latent-code optimization at test time, we train two 3D convolutional encoders $f_{\Omega_{\rs}}$ and $f_{\Omega_{\rp}}$ to predict initial estimates of the latent shape and pose codes, respectively.
Both encoders take as input the back-projected depth observation in the form of a partial voxel grid $V$.
We then employ 3D convolutions and a final fully-connected layer to output a latent code estimate.

\paragraph{Shape encoder.}
In particular, given the list of shape codes $\{ \fs_i \}_{i=1}^S$ learned from the $S$ identities in the train dataset, and a set of $P$ voxel grids of the available $P$ posed shapes in the training dataset $\{ V_j \}_{j=1}^P$, we train $f_{\Omega_{\rs}}$ to predict the mapping from the voxel grid $V_j$ to the corresponding shape code of the underlying identity.

\paragraph{Pose encoder.}
Similarly, given the list of pose codes $\{ \fp_j \}_{j=1}^P$ learned from the $P$ posed shapes in the dataset, and the set of $P$ voxel grids $\{ V_j \}_{j=1}^P$, we train $f_{\Omega_{\rp}}$ to predict the mapping from the voxel grid $V_j$ of a posed shape to the corresponding pose code.

\begin{figure}[t]
    \centering
    \fontsize{9pt}{11pt}\selectfont
    \def\svgwidth{\linewidth}
    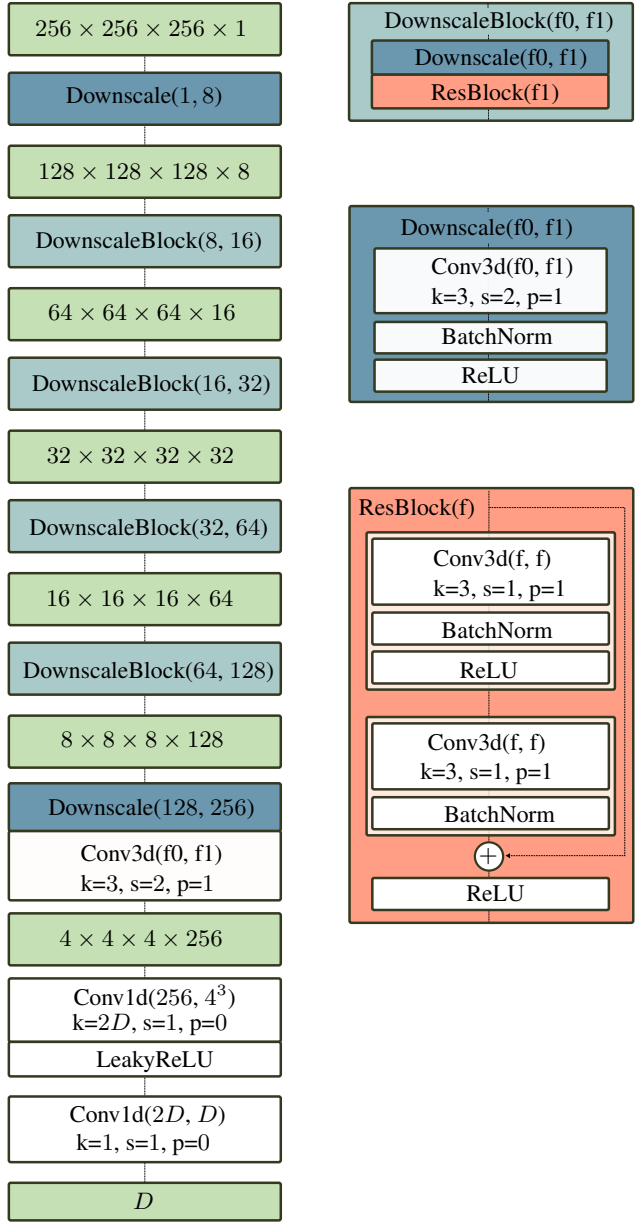
    \caption{
        A visualization of our 3D shape encoder for latent-code initialization at test time.
        Green blocks represent tensors, while the remaining blocks depict operations.
    }
    \label{fig:encoder}
\end{figure}

\vspace{0.5cm}
Fig.~\ref{fig:encoder} visualizes the encoder architecture employed for $f_{\Omega_{\rs}}$, with $D$ the output latent code dimension.
A similar architecture is employed to learn $f_{\Omega_{\rp}}$, differing only in the output channel dimension of the 3D convolution operations.
In particular, in our pose encoder we employ output channel dimensions of 16, 32, 64, 128, 256, 256, respectively for each 3D convolutional block in Fig~\ref{fig:encoder}.

\section{Training Details}
\label{sec:training_details}

We train both our shape and pose MLPs until convergence; in practice, this required $4000$ epochs for the shape MLP, and $150$ epochs for the pose MLP, which amounts to a similar number of iterations for each.
We use a batch size of $4$ in both cases.
Our human \OUR{} was trained on a GeForce RTX 3090 for approximately 8 days in total, which could be accelerated by parallelizing training on multiple GPUs.

\paragraph{Shape space.} For each identity in a mini-batch, out of the total $N_{\rs}$ samples available for a given identity in its canonical shape, we randomly sub-sample $50$k training points, 70\% of which are drawn from the available $N_{\rs}^{\mathrm{ns}}$ near-surface points, with the remaining 30\% drawn from the set of $N_{\rs}^{\mathrm{u}}$ uniform samples.

\paragraph{Pose space.} For each posed shape in a mini-batch, out of the total $N_{\rp}$ flow samples available, we randomly sub-sample $50$k correspondences.

\section{Inference-time Optimization Details}
\label{sec:inference_details}

When fitting an \OUR{} to an input monocular depth sequence of $L$ frames, we optimize for the shape code $\fs$ and $L$ pose codes $\{ \fp_j \}_{j=1}^{L}$ by minimizing Eq.~5 from the main text.
In our experiments, we optimize for a total of $I = 1000$ iterations.
We use the Adam optimizer~\cite{kingma2014adam} and learning rates of \SI{5e-4}{} and \SI{1e-3}{} for the shape and pose codes, respectively.
We decrease these learning rates by a factor of $0.5$ after every $250$ iterations, and control the temporal regularization loss $\mathcal{L}_{t}$ in Eq.~5 with a weight of $100$.

During test-time optimization, each element in a mini-batch consists of one frame from the sequence; we use a batch size of 4.
Before optimization begins, we make initial estimates of the shape code and the $L$ pose codes.
For the former, we use our shape encoder $f_{\Omega_{\rs}}$ to make shape code estimates for all $L$ frames in the input sequence.
We then compute the average vector and use this as initial shape code estimate.
As for the pose codes, we employ our pose encoder $f_{\Omega_{\rp}}$ to estimate a pose code for each frame in the sequence.

At this point, given the initial shape code estimate, we can extract an initial canonical shape by querying our learned shape MLP on a 3D grid, and then running Marching Cubes to extract the surface \cite{lorensen1987marching}, as explained in Sec.~3.2.
We then pre-sample \mbox{$N_t = \SI{500}{}$k} points, \mbox{$\{ \fx_k \}_{k=1}^{N_t}$}, around this initial estimate of the canonical shape; during optimization, for each frame in a mini-batch, we sub-sample \mbox{$N_b = 20$}k points --out of the available $N_t$-- to minimize Eq.~5.

As presented in the main text, we employ an additional ICP-like (Iterative Closest Point) term --denoted by $\mathcal{L}_{\mathrm{icp}}$ in Eq.~5-- that enforces surface points to be close to the input point cloud $Q_j$, which results from back-projecting the observed depth map.
In particular, at the beginning of every iteration, out of the $N_b$ points $\fx_k$ sampled in the canonical space for each element in a mini-batch, we select those points $\fx_k^{\mathrm{ns}}$ within a distance $\epsilon_{\mathrm{icp}}$ from the implicitly represented surface of the canonical shape, \ie, \mbox{$\fx_k^{\mathrm{ns}} = \fx_k \mid |f_{\theta_{\rs}}(\fs, \fx_k)| < \epsilon_{\mathrm{icp}}$} (in our experiments, \mbox{$\epsilon_{\mathrm{icp}} = 0.001$} in normalized coordinates).
Then, for every observed point $\bm{q} \in Q_j$, we minimize the distance to its nearest neighbor in the set of predicted model points in the $j$-th frame, denoted by $\mathcal{R} = \{ \fx_k^{\mathrm{ns}} + f_{\theta_{\rp}}(\fs, \fp_j, \fx_k^{\mathrm{ns}}) \}$:
\begin{equation}
    \mathcal{L}_{\mathrm{icp}} = \lambda_{\mathrm{icp}} \sum_{\bm{q} \in Q_j} \norm{ \bm{q} - \mathrm{NN}_{\mathcal{R}}(\bm{q}) }_2.
    \label{eq:icp_term}
\end{equation}
In the above equation, $\mathrm{NN}_{\mathcal{R}}(\cdot)$ denotes a function that queries the nearest neighbor of a 3D point in a set of points $\mathcal{R}$.
We control the importance of this loss with $\lambda_{\mathrm{icp}}$, which we set to $0.0005$ in our experiments.
We found this loss to be specially beneficial at the beginning of the optimization, in order to avoid falling in local minima. %
We then disable it, \ie, $\lambda_{\mathrm{icp}} = 0$, after $I/2$ iterations.
In Tab.~\ref{tab:ablation} we show the contribution of our ICP term.

\vspace{0.3cm}
As a reference, optimizing over an input sequence of 100 frames takes approximately 4 hours on a GeForce~RTX~3090 with our unoptimized implementation.

\section{Ablation Studies}
\label{sec:ablation}

Table~\ref{tab:ablation} shows an ablation study on the effect of training the shape and pose MLPs without positional encoding, on the latent code size, as well as on the effect of several test-time hyperparameters: not using any kind of temporal consistency, enforcing temporal consistency on the pose codes (instead of on the flow predictions), and, finally, on the effect of our ICP loss, as already mentioned in Sec.~\ref{sec:inference_details}.

\begin{table}[t]
	\resizebox{\linewidth}{!}{
    \centering
    \begin{tabular}{lcccc}
        \toprule
        \textbf{Method} & \textbf{IoU} $\uparrow$ & \textbf{C-$\ell_2$} ($\times 10^{-3}$) $\downarrow$ & \textbf{EPE} ($\times 10^{-2}$) $\downarrow$ \\ 
        \midrule
        Ours (w/o Positional Enc.) & 0.76 & 0.142 & 1.29 \\
        \midrule
        Ours (code size 128) & 0.81 & 0.052 & 0.95 \\
        Ours (code size 384) & 0.82 & 0.076 & 1.05 \\
        \midrule
        Ours (w/o temp. consist.) & 0.82 & 0.037 & 0.81 \\
        Ours (w/ code consist.) & 0.82 & 0.034 & 0.75 \\
        \midrule
        Ours (w/o ICP loss) & 0.83 & 0.031 & 0.79 \\
        \midrule
        Ours               & \textbf{0.83} & \textbf{0.022} & \textbf{0.74} \\
        \bottomrule
    \end{tabular}
    }
    \caption{
        Ablation study on CAPE \cite{CAPE:CVPR:20} to evaluate the contribution of our different modules and network choices.
    }
    \label{tab:ablation}
\end{table}

\section{Robustness to Initialization}
\label{sec:robustness}

\begin{figure}[t]
    \centering
    \fontsize{7pt}{11pt}\selectfont
    \def\svgwidth{\linewidth}
\begingroup%
  \makeatletter%
  \providecommand\color[2][]{%
    \errmessage{(Inkscape) Color is used for the text in Inkscape, but the package 'color.sty' is not loaded}%
    \renewcommand\color[2][]{}%
  }%
  \providecommand\transparent[1]{%
    \errmessage{(Inkscape) Transparency is used (non-zero) for the text in Inkscape, but the package 'transparent.sty' is not loaded}%
    \renewcommand\transparent[1]{}%
  }%
  \providecommand\rotatebox[2]{#2}%
  \newcommand*\fsize{\dimexpr\f@size pt\relax}%
  \newcommand*\lineheight[1]{\fontsize{\fsize}{#1\fsize}\selectfont}%
  \ifx\svgwidth\undefined%
    \setlength{\unitlength}{299.40273589bp}%
    \ifx\svgscale\undefined%
      \relax%
    \else%
      \setlength{\unitlength}{\unitlength * \real{\svgscale}}%
    \fi%
  \else%
    \setlength{\unitlength}{\svgwidth}%
  \fi%
  \global\let\svgwidth\undefined%
  \global\let\svgscale\undefined%
  \makeatother%
  \begin{picture}(1,0.52280352)%
    \lineheight{1}%
    \setlength\tabcolsep{0pt}%
    \put(0,0){\includegraphics[width=\unitlength,page=1]{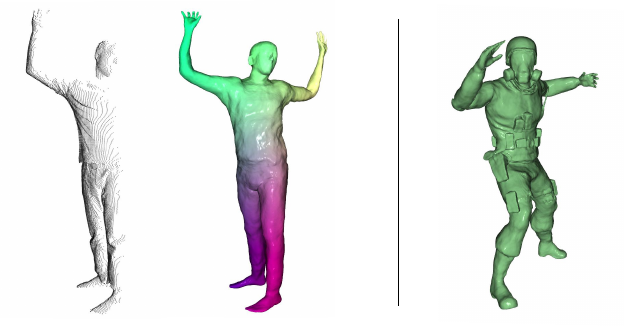}}%
    \put(0.26008753,0.01976753){\color[rgb]{0,0,0}\makebox(0,0)[lt]{\lineheight{1.25}\smash{\begin{tabular}[t]{l}(a)\end{tabular}}}}%
    \put(0.86717465,0.01976753){\color[rgb]{0,0,0}\makebox(0,0)[lt]{\lineheight{1.25}\smash{\begin{tabular}[t]{l}(b)\end{tabular}}}}%
  \end{picture}%
\endgroup%

    \caption{
        (a) Robustness in the presence of strong occlusions. (b) Example of complex topology and clothing.
    }
    \label{fig:robustness_and_additional_examples}
\end{figure}

Latent code initialization via learned encoders offers additional robustness against local minima, but even without such initial code estimates, our method  achieves competitive performance (Table~1 in the main text).
Additionally, methods such as SMPL rely on strong initialization, which we provide with human joint predictions computed with OpenPose.
Without this additional guidance, optimizing over SMPL parameters becomes ill-posed, particularly for the task of monocular depth fitting.
In contrast, our proposed model fitting is robust to self-occlusions caused by rotations (Fig.~\ref{fig:robustness_and_additional_examples}a). 

\section{Consistency of Shape Fitting}
\label{sec:consistency}

To evaluate the consistency of shape fitting in \OURS{}, we reconstructed two consecutive sub-sequences of 50 frames each, and measured the similarity between the corresponding shape estimates, which matched almost perfectly ($0.96$~IoU and $1.61 \times 10^{-6}$~chamfer).

\section{Additional Examples}
\label{sec:additional}

To further showcase the expressiveness and potential of \OURS{}, in Figure \ref{fig:robustness_and_additional_examples}b we show an additional identity drawn from the latent space featuring complex topologies and clothing.

\section{Comparison with Zhou et al. \cite{zhou2020unsupervised}}
\label{sec:comparison_unsup}
\begin{figure}[t]
    \centering
    \fontsize{7pt}{11pt}\selectfont
    \def\svgwidth{\linewidth}
\begingroup%
  \makeatletter%
  \providecommand\color[2][]{%
    \errmessage{(Inkscape) Color is used for the text in Inkscape, but the package 'color.sty' is not loaded}%
    \renewcommand\color[2][]{}%
  }%
  \providecommand\transparent[1]{%
    \errmessage{(Inkscape) Transparency is used (non-zero) for the text in Inkscape, but the package 'transparent.sty' is not loaded}%
    \renewcommand\transparent[1]{}%
  }%
  \providecommand\rotatebox[2]{#2}%
  \newcommand*\fsize{\dimexpr\f@size pt\relax}%
  \newcommand*\lineheight[1]{\fontsize{\fsize}{#1\fsize}\selectfont}%
  \ifx\svgwidth\undefined%
    \setlength{\unitlength}{248.61794853bp}%
    \ifx\svgscale\undefined%
      \relax%
    \else%
      \setlength{\unitlength}{\unitlength * \real{\svgscale}}%
    \fi%
  \else%
    \setlength{\unitlength}{\svgwidth}%
  \fi%
  \global\let\svgwidth\undefined%
  \global\let\svgscale\undefined%
  \makeatother%
  \begin{picture}(1,0.66531557)%
    \lineheight{1}%
    \setlength\tabcolsep{0pt}%
    \put(0,0){\includegraphics[width=\unitlength,page=1]{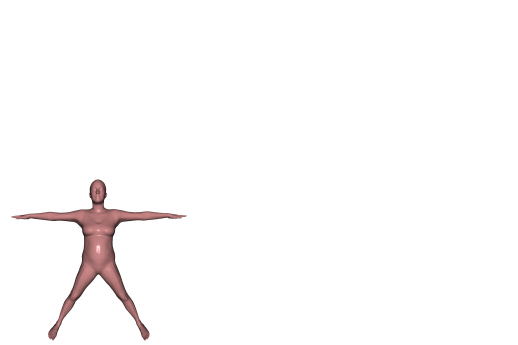}}%
    \put(0.01701414,0.36542225){\color[rgb]{0,0,0}\rotatebox{90}{\makebox(0,0)[lt]{\lineheight{1.25}\smash{\begin{tabular}[t]{l}Pose Source\end{tabular}}}}}%
    \put(0.02077034,0.02280061){\color[rgb]{0,0,0}\rotatebox{90}{\makebox(0,0)[lt]{\lineheight{1.25}\smash{\begin{tabular}[t]{l}Shape Source\end{tabular}}}}}%
    \put(0,0){\includegraphics[width=\unitlength,page=2]{comparison_unsup.pdf}}%
    \put(0.39674313,0.10617578){\color[rgb]{0,0,0}\makebox(0,0)[lt]{\lineheight{1.25}\smash{\begin{tabular}[t]{l}\cite{zhou2020unsupervised}\end{tabular}}}}%
    \put(0.73336629,0.10582331){\color[rgb]{0,0,0}\makebox(0,0)[lt]{\lineheight{1.25}\smash{\begin{tabular}[t]{l}Ours\end{tabular}}}}%
    \put(0,0){\includegraphics[width=\unitlength,page=3]{comparison_unsup.pdf}}%
  \end{picture}%
\endgroup%

    \caption{
        Comparison with \cite{zhou2020unsupervised} on the task of pose transfer.
    }
    \label{fig:comparison_unsup}
\end{figure}

Additionally, we compare with \cite{zhou2020unsupervised} on the task of pose transfer.
In contrast to \cite{zhou2020unsupervised}, our focus is to construct disentangled spaces which enable test-time fitting to new observations.
Furthermore, \cite{zhou2020unsupervised} uses a mesh auto-encoder, limiting the approach to only representing a fixed topology. 
Our approach naturally deals with complex topology, as we have shown in Sec.~4 in the main paper.
For this experiment, we trained an \OUR{} on the same human body dataset as \cite{zhou2020unsupervised}, and show a comparison in Fig.~\ref{fig:comparison_unsup}.

\end{appendix}


{\small
\bibliographystyle{ieee_fullname}
\bibliography{main}

\begin{thebibliography}{10}\itemsep=-1pt

\bibitem{alldieck2019learning}
Thiemo Alldieck, Marcus Magnor, Bharat~Lal Bhatnagar, Christian Theobalt, and
  Gerard Pons-Moll.
\newblock Learning to reconstruct people in clothing from a single rgb camera.
\newblock In {\em Proceedings of the IEEE/CVF Conference on Computer Vision and
  Pattern Recognition}, pages 1175--1186, 2019.

\bibitem{anguelov2005scape}
Dragomir Anguelov, Praveen Srinivasan, Daphne Koller, Sebastian Thrun, Jim
  Rodgers, and James Davis.
\newblock Scape: shape completion and animation of people.
\newblock In {\em ACM SIGGRAPH 2005 Papers}, pages 408--416. 2005.

\bibitem{bhatnagar2020ipnet}
Bharat~Lal Bhatnagar, Cristian Sminchisescu, Christian Theobalt, and Gerard
  Pons-Moll.
\newblock Combining implicit function learning and parametric models for 3d
  human reconstruction.
\newblock {\em arXiv preprint arXiv:2007.11432}, 2020.

\bibitem{bhatnagar2019multi}
Bharat~Lal Bhatnagar, Garvita Tiwari, Christian Theobalt, and Gerard Pons-Moll.
\newblock Multi-garment net: Learning to dress 3d people from images.
\newblock In {\em Proceedings of the IEEE/CVF International Conference on
  Computer Vision}, pages 5420--5430, 2019.

\bibitem{bogo2017dynamicFAUST}
Federica Bogo, Javier Romero, Gerard Pons-Moll, and Michael~J Black.
\newblock Dynamic faust: Registering human bodies in motion.
\newblock In {\em Proceedings of the IEEE conference on computer vision and
  pattern recognition}, pages 6233--6242, 2017.

\bibitem{bovzivc2020neural}
Alja{\v{z}} Bo{\v{z}}i{\v{c}}, Pablo Palafox, Michael Zollh{\"o}fer, Justus
  Thies, Angela Dai, and Matthias Nie{\ss}ner.
\newblock Neural deformation graphs for globally-consistent non-rigid
  reconstruction.
\newblock {\em arXiv preprint arXiv:2012.01451}, 2020.

\bibitem{bozic2020neuraltracking}
Alja{\v{z}} Bo{\v{z}}i{\v{c}}, Pablo Palafox, Michael Zoll{\"o}fer, Angela Dai,
  Justus Thies, and Matthias Nie{\ss}ner.
\newblock Neural non-rigid tracking.
\newblock In {\em NeurIPS}, 2020.

\bibitem{bozic2020deepdeform}
Aljaz Bozic, Michael Zollhofer, Christian Theobalt, and Matthias Nie{\ss}ner.
\newblock Deepdeform: Learning non-rigid rgb-d reconstruction with
  semi-supervised data.
\newblock In {\em Proceedings of the IEEE/CVF Conference on Computer Vision and
  Pattern Recognition}, pages 7002--7012, 2020.

\bibitem{cao2019openpose}
Zhe Cao, Gines Hidalgo, Tomas Simon, Shih-En Wei, and Yaser Sheikh.
\newblock Openpose: realtime multi-person 2d pose estimation using part
  affinity fields.
\newblock {\em IEEE transactions on pattern analysis and machine intelligence},
  43(1):172--186, 2019.

\bibitem{chen2019learning}
Zhiqin Chen and Hao Zhang.
\newblock Learning implicit fields for generative shape modeling.
\newblock In {\em Proceedings of the IEEE/CVF Conference on Computer Vision and
  Pattern Recognition}, pages 5939--5948, 2019.

\bibitem{chibane2020implicit}
Julian Chibane, Thiemo Alldieck, and Gerard Pons-Moll.
\newblock Implicit functions in feature space for 3d shape reconstruction and
  completion.
\newblock In {\em Proceedings of the IEEE/CVF Conference on Computer Vision and
  Pattern Recognition}, pages 6970--6981, 2020.

\bibitem{choy20163d}
Christopher~B Choy, Danfei Xu, JunYoung Gwak, Kevin Chen, and Silvio Savarese.
\newblock 3d-r2n2: A unified approach for single and multi-view 3d object
  reconstruction.
\newblock In {\em European conference on computer vision}, pages 628--644.
  Springer, 2016.

\bibitem{dai2017bundlefusion}
Angela Dai, Matthias Nie{\ss}ner, Michael Zollh{\"{o}}fer, Shahram Izadi, and
  Christian Theobalt.
\newblock Bundlefusion: Real-time globally consistent 3d reconstruction using
  on-the-fly surface reintegration.
\newblock {\em {ACM} Trans. Graph.}, 36(3):24:1--24:18, 2017.

\bibitem{dai2017complete}
Angela Dai, Charles~Ruizhongtai Qi, and Matthias Nie{\ss}ner.
\newblock Shape completion using 3d-encoder-predictor cnns and shape synthesis.
\newblock In {\em 2017 {IEEE} Conference on Computer Vision and Pattern
  Recognition, {CVPR} 2017, Honolulu, HI, USA, July 21-26, 2017}, pages
  6545--6554, 2017.

\bibitem{dai2017shape}
Angela Dai, Charles Ruizhongtai~Qi, and Matthias Nie{\ss}ner.
\newblock Shape completion using 3d-encoder-predictor cnns and shape synthesis.
\newblock In {\em Proceedings of the IEEE Conference on Computer Vision and
  Pattern Recognition}, pages 5868--5877, 2017.

\bibitem{fan2017point}
Haoqiang Fan, Hao Su, and Leonidas~J Guibas.
\newblock A point set generation network for 3d object reconstruction from a
  single image.
\newblock In {\em Proceedings of the IEEE conference on computer vision and
  pattern recognition}, pages 605--613, 2017.

\bibitem{genova2019learning}
Kyle Genova, Forrester Cole, Daniel Vlasic, Aaron Sarna, William~T Freeman, and
  Thomas Funkhouser.
\newblock Learning shape templates with structured implicit functions.
\newblock In {\em Proceedings of the IEEE/CVF International Conference on
  Computer Vision}, pages 7154--7164, 2019.

\bibitem{groueix20183d}
Thibault Groueix, Matthew Fisher, Vladimir~G Kim, Bryan~C Russell, and Mathieu
  Aubry.
\newblock 3d-coded: 3d correspondences by deep deformation.
\newblock In {\em Proceedings of the European Conference on Computer Vision
  (ECCV)}, pages 230--246, 2018.

\bibitem{izadi2011kinectfusion}
Shahram Izadi, David Kim, Otmar Hilliges, David Molyneaux, Richard~A. Newcombe,
  Pushmeet Kohli, Jamie Shotton, Steve Hodges, Dustin Freeman, Andrew~J.
  Davison, and Andrew~W. Fitzgibbon.
\newblock Kinectfusion: real-time 3d reconstruction and interaction using a
  moving depth camera.
\newblock In {\em Proceedings of the 24th Annual {ACM} Symposium on User
  Interface Software and Technology, Santa Barbara, CA, USA, October 16-19,
  2011}, pages 559--568, 2011.

\bibitem{joo2018total}
Hanbyul Joo, Tomas Simon, and Yaser Sheikh.
\newblock Total capture: A 3d deformation model for tracking faces, hands, and
  bodies.
\newblock In {\em Proceedings of the IEEE conference on computer vision and
  pattern recognition}, pages 8320--8329, 2018.

\bibitem{kazhdan2006poisson}
Michael Kazhdan, Matthew Bolitho, and Hugues Hoppe.
\newblock Poisson surface reconstruction.
\newblock In {\em Proceedings of the fourth Eurographics symposium on Geometry
  processing}, volume~7, 2006.

\bibitem{kingma2014adam}
Diederik~P Kingma and Jimmy Ba.
\newblock Adam: A method for stochastic optimization.
\newblock {\em arXiv preprint arXiv:1412.6980}, 2014.

\bibitem{lazova2019360}
Verica Lazova, Eldar Insafutdinov, and Gerard Pons-Moll.
\newblock 360-degree textures of people in clothing from a single image.
\newblock In {\em 2019 International Conference on 3D Vision (3DV)}, pages
  643--653. IEEE, 2019.

\bibitem{li2017learning_flame}
Tianye Li, Timo Bolkart, Michael~J Black, Hao Li, and Javier Romero.
\newblock Learning a model of facial shape and expression from 4d scans.
\newblock {\em ACM Trans. Graph.}, 36(6):194--1, 2017.

\bibitem{li2020learning}
Yang Li, Aljaz Bozic, Tianwei Zhang, Yanli Ji, Tatsuya Harada, and Matthias
  Nie{\ss}ner.
\newblock Learning to optimize non-rigid tracking.
\newblock In {\em Proceedings of the IEEE/CVF Conference on Computer Vision and
  Pattern Recognition}, pages 4910--4918, 2020.

\bibitem{yang20204dcomplete}
Yang Li, Hikari Takehara, Takafumi Taketomi, Bo Zheng, and Matthias
  Nie{\ss}ner.
\newblock 4dcomplete: Non-rigid motion estimation beyond the observable
  surface.

\bibitem{loper2015smpl}
Matthew Loper, Naureen Mahmood, Javier Romero, Gerard Pons-Moll, and Michael~J
  Black.
\newblock Smpl: A skinned multi-person linear model.
\newblock {\em ACM transactions on graphics (TOG)}, 34(6):1--16, 2015.

\bibitem{lorensen1987marching}
William~E Lorensen and Harvey~E Cline.
\newblock Marching cubes: A high resolution 3d surface construction algorithm.
\newblock {\em ACM siggraph computer graphics}, 21(4):163--169, 1987.

\bibitem{ma2020learning}
Qianli Ma, Jinlong Yang, Anurag Ranjan, Sergi Pujades, Gerard Pons-Moll, Siyu
  Tang, and Michael~J Black.
\newblock Learning to dress 3d people in generative clothing.
\newblock In {\em Proceedings of the IEEE/CVF Conference on Computer Vision and
  Pattern Recognition}, pages 6469--6478, 2020.

\bibitem{CAPE:CVPR:20}
Qianli Ma, Jinlong Yang, Anurag Ranjan, Sergi Pujades, Gerard Pons-Moll, Siyu
  Tang, and Michael~J. Black.
\newblock Learning to dress 3d people in generative clothing.
\newblock In {\em Computer Vision and Pattern Recognition (CVPR)}, June 2020.

\bibitem{AMASS:ICCV:2019}
Naureen Mahmood, Nima Ghorbani, Nikolaus~F. Troje, Gerard Pons-Moll, and
  Michael~J. Black.
\newblock {AMASS}: Archive of motion capture as surface shapes.
\newblock In {\em International Conference on Computer Vision}, pages
  5442--5451, Oct. 2019.

\bibitem{mescheder2019occupancyNet}
Lars Mescheder, Michael Oechsle, Michael Niemeyer, Sebastian Nowozin, and
  Andreas Geiger.
\newblock Occupancy networks: Learning 3d reconstruction in function space.
\newblock In {\em Proceedings of the IEEE Conference on Computer Vision and
  Pattern Recognition}, pages 4460--4470, 2019.

\bibitem{michalkiewicz2019deep}
Mateusz Michalkiewicz, Jhony~K Pontes, Dominic Jack, Mahsa Baktashmotlagh, and
  Anders Eriksson.
\newblock Deep level sets: Implicit surface representations for 3d shape
  inference.
\newblock {\em arXiv preprint arXiv:1901.06802}, 2019.

\bibitem{mildenhall2020nerf}
Ben Mildenhall, Pratul~P Srinivasan, Matthew Tancik, Jonathan~T Barron, Ravi
  Ramamoorthi, and Ren Ng.
\newblock Nerf: Representing scenes as neural radiance fields for view
  synthesis.
\newblock In {\em European Conference on Computer Vision}, pages 405--421.
  Springer, 2020.

\bibitem{newcombe2015dynamicfusion}
Richard~A Newcombe, Dieter Fox, and Steven~M Seitz.
\newblock Dynamicfusion: Reconstruction and tracking of non-rigid scenes in
  real-time.
\newblock In {\em Proceedings of the IEEE Conference on Computer Vision and
  Pattern Recognition (CVPR)}, pages 343--352, 2015.

\bibitem{newcombe2011kinectfusion}
Richard~A. Newcombe, Shahram Izadi, Otmar Hilliges, David Molyneaux, David Kim,
  Andrew~J. Davison, Pushmeet Kohli, Jamie Shotton, Steve Hodges, and Andrew~W.
  Fitzgibbon.
\newblock Kinectfusion: Real-time dense surface mapping and tracking.
\newblock In {\em 10th {IEEE} International Symposium on Mixed and Augmented
  Reality, {ISMAR} 2011, Basel, Switzerland, October 26-29, 2011}, pages
  127--136, 2011.

\bibitem{niemeyer2019occupancyFlow}
Michael Niemeyer, Lars Mescheder, Michael Oechsle, and Andreas Geiger.
\newblock Occupancy flow: 4d reconstruction by learning particle dynamics.
\newblock In {\em Proceedings of the IEEE International Conference on Computer
  Vision}, pages 5379--5389, 2019.

\bibitem{niessner2013hashing}
M. Nie{\ss}ner, M. Zollh\"ofer, S. Izadi, and M. Stamminger.
\newblock Real-time 3d reconstruction at scale using voxel hashing.
\newblock {\em ACM Transactions on Graphics (TOG)}, 2013.

\bibitem{park2019deepsdf}
Jeong~Joon Park, Peter Florence, Julian Straub, Richard Newcombe, and Steven
  Lovegrove.
\newblock Deepsdf: Learning continuous signed distance functions for shape
  representation.
\newblock In {\em Proceedings of the IEEE Conference on Computer Vision and
  Pattern Recognition}, pages 165--174, 2019.

\bibitem{patel2020virtual}
Chaitanya Patel, Zhouyingcheng Liao, and Gerard Pons-Moll.
\newblock The virtual tailor: Predicting clothing in 3d as a function of human
  pose, shape and garment style.
\newblock {\em arXiv preprint arXiv:2003.04583}, 2020.

\bibitem{paysan20093d}
Pascal Paysan, Reinhard Knothe, Brian Amberg, Sami Romdhani, and Thomas Vetter.
\newblock A 3d face model for pose and illumination invariant face recognition.
\newblock In {\em 2009 sixth IEEE international conference on advanced video
  and signal based surveillance}, pages 296--301. Ieee, 2009.

\bibitem{ploumpis2019combining}
Stylianos Ploumpis, Haoyang Wang, Nick Pears, William~AP Smith, and Stefanos
  Zafeiriou.
\newblock Combining 3d morphable models: A large scale face-and-head model.
\newblock In {\em Proceedings of the IEEE/CVF Conference on Computer Vision and
  Pattern Recognition}, pages 10934--10943, 2019.

\bibitem{pons2017clothcap}
Gerard Pons-Moll, Sergi Pujades, Sonny Hu, and Michael~J Black.
\newblock Clothcap: Seamless 4d clothing capture and retargeting.
\newblock {\em ACM Transactions on Graphics (TOG)}, 36(4):1--15, 2017.

\bibitem{pons2015dyna}
Gerard Pons-Moll, Javier Romero, Naureen Mahmood, and Michael~J Black.
\newblock Dyna: A model of dynamic human shape in motion.
\newblock {\em ACM Transactions on Graphics (TOG)}, 34(4):1--14, 2015.

\bibitem{MANO:SIGGRAPHASIA:2017}
Javier Romero, Dimitrios Tzionas, and Michael~J. Black.
\newblock Embodied hands: Modeling and capturing hands and bodies together.
\newblock {\em ACM Transactions on Graphics, (Proc. SIGGRAPH Asia)}, 36(6),
  Nov. 2017.

\bibitem{salimans2016weight}
Tim Salimans and Diederik~P Kingma.
\newblock Weight normalization: A simple reparameterization to accelerate
  training of deep neural networks.
\newblock {\em arXiv preprint arXiv:1602.07868}, 2016.

\bibitem{slavcheva2017killingfusion}
Miroslava Slavcheva, Maximilian Baust, Daniel Cremers, and Slobodan Ilic.
\newblock Killingfusion: Non-rigid 3d reconstruction without correspondences.
\newblock In {\em Proceedings of the IEEE Conference on Computer Vision and
  Pattern Recognition (CVPR)}, pages 1386--1395, 2017.

\bibitem{tiwari2020sizer}
Garvita Tiwari, Bharat~Lal Bhatnagar, Tony Tung, and Gerard Pons-Moll.
\newblock Sizer: A dataset and model for parsing 3d clothing and learning size
  sensitive 3d clothing.
\newblock {\em arXiv preprint arXiv:2007.11610}, 2020.

\bibitem{wang2018pixel2mesh}
Nanyang Wang, Yinda Zhang, Zhuwen Li, Yanwei Fu, Wei Liu, and Yu-Gang Jiang.
\newblock Pixel2mesh: Generating 3d mesh models from single rgb images.
\newblock In {\em Proceedings of the European Conference on Computer Vision
  (ECCV)}, pages 52--67, 2018.

\bibitem{wu20153d}
Zhirong Wu, Shuran Song, Aditya Khosla, Fisher Yu, Linguang Zhang, Xiaoou Tang,
  and Jianxiong Xiao.
\newblock 3d shapenets: {A} deep representation for volumetric shapes.
\newblock In {\em {IEEE} Conference on Computer Vision and Pattern Recognition,
  {CVPR} 2015, Boston, MA, USA, June 7-12, 2015}, pages 1912--1920, 2015.

\bibitem{xu2020ghum}
Hongyi Xu, Eduard~Gabriel Bazavan, Andrei Zanfir, William~T Freeman, Rahul
  Sukthankar, and Cristian Sminchisescu.
\newblock Ghum \& ghuml: Generative 3d human shape and articulated pose models.
\newblock In {\em Proceedings of the IEEE/CVF Conference on Computer Vision and
  Pattern Recognition}, pages 6184--6193, 2020.

\bibitem{zhou2020unsupervised}
Keyang Zhou, Bharat~Lal Bhatnagar, and Gerard Pons-Moll.
\newblock Unsupervised shape and pose disentanglement for 3d meshes.
\newblock In {\em European Conference on Computer Vision}, pages 341--357.
  Springer, 2020.

\bibitem{Zuffi_CVPR_2017}
Silvia Zuffi, Angjoo Kanazawa, David Jacobs, and Michael~J. Black.
\newblock {3D} menagerie: Modeling the {3D} shape and pose of animals.
\newblock In {\em IEEE Conf. on Computer Vision and Pattern Recognition
  (CVPR)}, July 2017.

\end{thebibliography}
}

\end{document}